\documentclass[letterpaper]{article} 
\usepackage[draft]{aaai2026}  
\usepackage{times}  
\usepackage{helvet}  
\usepackage{courier}  
\usepackage[hyphens]{url}  
\usepackage{graphicx} 
\usepackage{natbib}  
\usepackage{caption} 
\usepackage{amsmath}
\usepackage{amssymb}
\usepackage{booktabs}
\usepackage{algorithm}
\usepackage{algorithmic}
\usepackage{arydshln}
\usepackage{pifont}
\usepackage{multirow}
\usepackage{color, colortbl}
\usepackage{tcolorbox}
\usepackage{mdframed}
\usepackage{cleveref} 
\usepackage{pifont}       
\usepackage{xcolor}       
\usepackage{enumitem}     

\newcommand{\circledone}{\textcolor{blue}{\raisebox{0pt}{\Large\textcircled{\small 1}}}}
\newcommand{\circledtwo}{\textcolor{blue}{\raisebox{0pt}{\Large\textcircled{\small 2}}}}
\newcommand{\circledthree}{\textcolor{blue}{\raisebox{0pt}{\Large\textcircled{\small 3}}}}

\newcommand{\highlight}[1]{\textcolor{blue!70!black}{\textbf{#1}}}
\newcommand{\cmark}{\ding{51}} 
\newcommand{\xmark}{\ding{55}} 

\usepackage{newfloat}
\usepackage{listings}
\DeclareCaptionStyle{ruled}{labelfont=normalfont,labelsep=colon,strut=off} 
\floatstyle{ruled}
\newfloat{listing}{tb}{lst}{}
\floatname{listing}{Listing}
\title{DOFA-CLIP: Multimodal Vision--Language Foundation Models for \\ Earth Observation}

\author {
    Zhitong Xiong\textsuperscript{1} \qquad
    Yi Wang\textsuperscript{1} \qquad
    Weikang Yu\textsuperscript{1,2} \qquad
    Adam J Stewart\textsuperscript{1} \qquad
    Jie Zhao\textsuperscript{1} \\
    Nils Lehmann\textsuperscript{1} \qquad
    Thomas Dujardin\textsuperscript{1} \qquad
    Zhenghang Yuan\textsuperscript{1} \qquad
    Pedram Ghamisi\textsuperscript{2} \\ 
    Xiao Xiang Zhu~\textsuperscript{1*} \\
    \textsuperscript{1} Technical University of Munich \qquad \textsuperscript{2} Helmholtz-Zentrum Dresden-Rossendorf \\
    {\tt\small \{zhitong.xiong, xiaoxiang.zhu\}@tum.de}
}

\usepackage{bibentry}

\begin{document}

\maketitle

\begin{abstract}
Earth observation (EO) spans a broad spectrum of modalities, including optical, radar, multispectral, and hyperspectral data, each capturing distinct environmental signals. However, current vision-language models in EO, particularly CLIP-based variants, remain confined to individual modalities, limiting generalization and scalability across diverse tasks. We present DOFA-CLIP (Dynamic-One-For-All CLIP), a unified vision-language foundation model that dynamically adapts to EO modalities with flexible spectral configurations through a single Transformer backbone. Our approach introduces three key contributions: 1) the construction of GeoLangBind-2M, a large-scale EO image–text dataset covering six heterogeneous modalities with rich natural language descriptions; 2) a novel training strategy called VECT (Vision-models Enhanced Contrastive Text-image pretraining), which enhances the spatial awareness of CLIP features with multiple vision foundation models; and 3) a Modality-aware Knowledge Agglomeration (MaKA) module that refines feature distillation with modality-specific awareness. DOFA-CLIP achieves state-of-the-art zero-shot performance across a wide range of EO benchmarks, including unseen modalities and a diverse number of input spectral bands. Together, these contributions establish a scalable foundation for multimodal EO understanding and open new avenues for integrating heterogeneous EO data with large language models. Code and datasets will be released.
\end{abstract}

\section{Introduction}

\begin{figure}[tp]
    \centering
    \includegraphics[width=\linewidth]{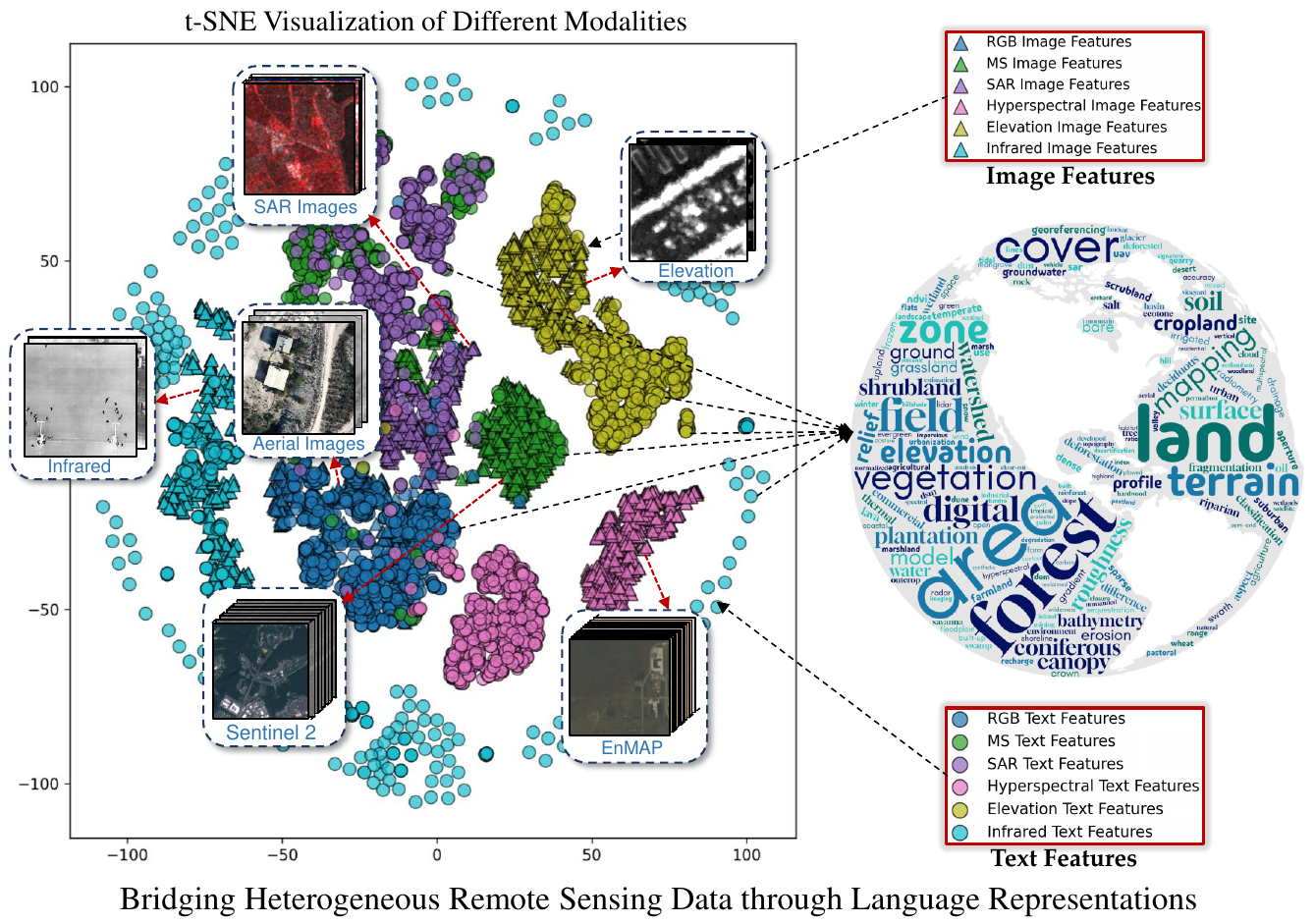}
    \caption{DOFA-CLIP employs a single unified vision Transformer to align heterogeneous EO data modalities using language as the grounding signal for cross-modal understanding and zero-shot generalization.}
    \label{fig:main_idea}
\end{figure}

Earth observation (EO) technologies have undergone remarkable advancement in recent years, generating vast amounts of multimodal geospatial data spanning different sensors and scales~\cite{xia2018dota,zhu2017deep,xiong2024earthnets}. In response to the exponential growth of remote sensing data, researchers have proposed foundation models~\cite{cong2022satmae,reed2023scale,wang2024hypersigma,wang2023ssl4eo,stewart2023ssl4eo,hong2024spectralgpt} pretrained on large-scale EO data for multiple downstream tasks. However, these models typically focus on individual EO data modality, limiting their ability to leverage complementary characteristics across different sensor types~\cite{zhu2024foundations}.

To address this limitation, recent works such as OFA-Net~\cite{OFA}, DOFA~\cite{xiong2024neural}, AnySat~\cite{astruc2024anysat}, OmniSat~\cite{astruc2024omnisat}, and FoMo-Net~\cite{bountos2023fomo} have introduced foundation models capable of learning from multiple EO modalities within a unified framework. These approaches have demonstrated promising results in multimodal EO representation learning. However, their focus remains on unsupervised visual pretraining, lacking a structured way to align EO data with language for better interpretability and retrieval. Meanwhile, multimodal large language models like GPT~\cite{openai2023gpt4}, LLaVA~\cite{liu2023visual}, and Pixtral~\cite{agrawal2024pixtral} have shown remarkable capabilities in handling diverse and complex tasks. Vision-language foundation models, such as CLIP~\cite{radford2021learning} and SigLIP~\cite{zhai2023sigmoid}, have played a crucial role as vision encoders of such multimodal large language models. Inspired by these advances, EO-specific adaptations like RemoteCLIP~\cite{liu2024remoteclip}, RS5M~\cite{zhang2024rs5m}, RS-CLIP~\cite{li2023rs}, and SkyScript~\cite{wang2024skyscript} have attempted to bring vision-language contrastive learning to remote sensing. However, existing CLIP models for EO have limitations: 1) These models are usually limited to high-resolution RGB imagery and cannot handle modalities like multi/hyper-spectral or SAR, which are vital for many applications. 2) Compared with vision models like DINOv2~\cite{oquab2023dinov2}, CLIP models are unsuitable for dense understanding tasks due to the lack of spatial awareness.

Training a single powerful vision–language model capable of handling heterogeneous EO data modalities raises several core research questions (\highlight{RQ}). These include:
\begin{itemize}[leftmargin=2em, itemsep=0.1em]
    \item[\circledone] \highlight{RQ1: How can we build a large-scale, semantically rich image--text dataset for diverse heterogeneous EO modalities?}
    \item[\circledtwo] \highlight{RQ2: How can we efficiently train large CLIP models for EO with better spatial awareness?}
    \item[\circledthree] \highlight{RQ3: How can a single CLIP model handle heterogeneous EO data while preserving modality-specific characteristics?}
\end{itemize}

To address such challenges, we propose Dynamic-One-For-All CLIP (DOFA-CLIP), a unified multimodal vision–language foundation model that aligns diverse EO modalities to the language space using a single Transformer, as shown in Fig. \ref{fig:main_idea}. Our contributions are as follows:
\begin{enumerate}
    \item We introduce GeoLangBind-2M, a large-scale EO dataset comprising two million image–text pairs across six heterogeneous modalities, providing high-quality text descriptions for multimodal vision–language training.
    \item We propose Vision-models Enhanced Contrastive Text-image pretraining (VECT), a simple yet effective training strategy that enhances the fine-grained spatial awareness of CLIP models with open vision foundation models.
    \item We develop the Modality-aware Knowledge Agglomeration (MaKA) module to support modality-adaptive representation distillation within a unified Transformer.
    \item We conduct extensive evaluations across zero-shot classification and cross-modal retrieval tasks, demonstrating state-of-the-art performance and better spatial awareness on both seen and unseen EO modalities.
    \end{enumerate}

\begin{figure*}[tp]
    \centering
    \includegraphics[width=0.85\linewidth]{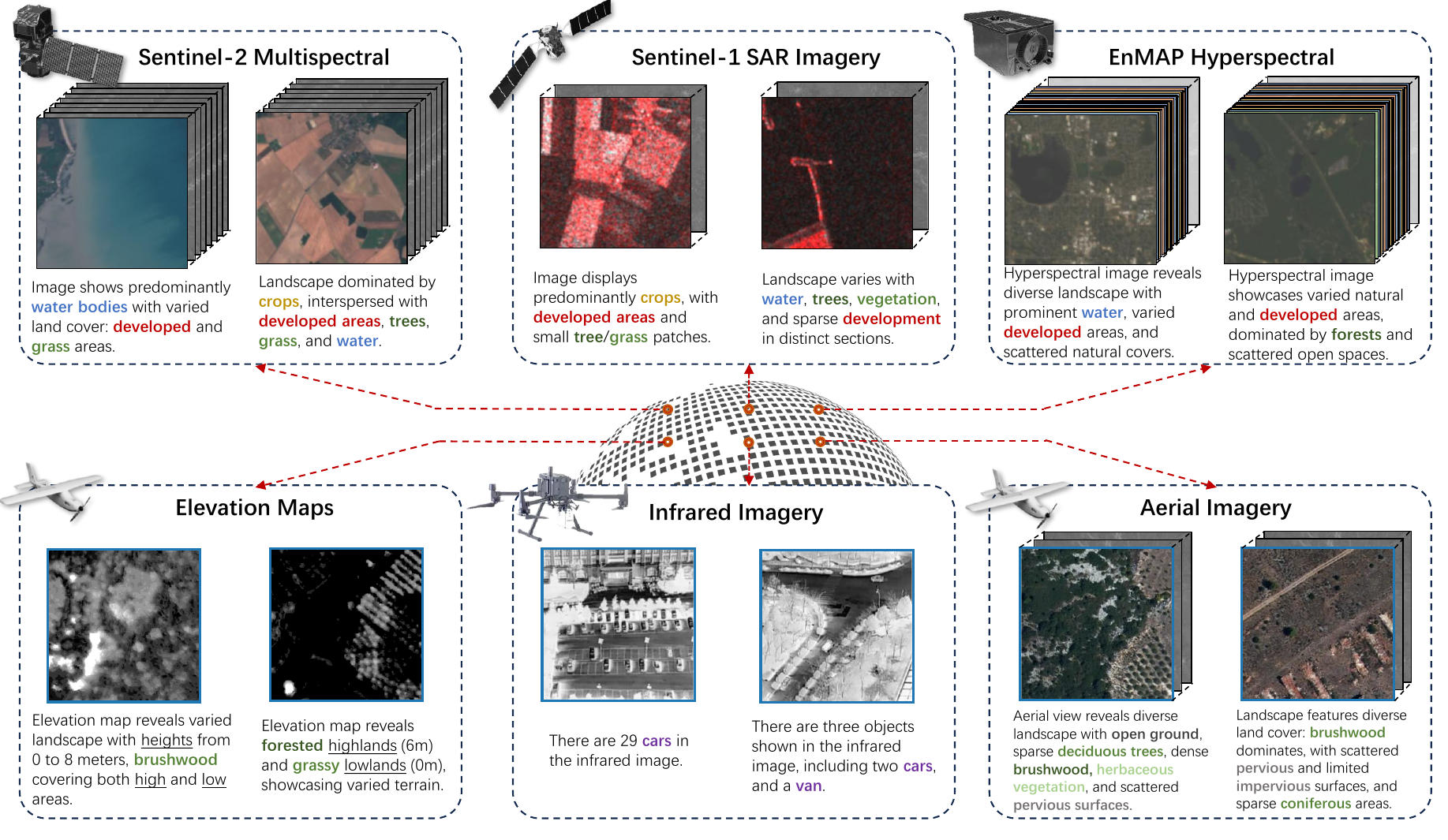}
    \caption{Visualization of data samples from GeoLangBind-2M. The dataset includes imagery from six different sensors and modalities: Sentinel-2 multispectral, Sentinel-1 SAR, EnMAP hyperspectral, elevation maps, infrared imagery, and aerial imagery. Each sample is paired with textual descriptions capturing key land cover types, objects, and geographic features.}
    \label{fig:vis_examples}
\end{figure*}

\section{Related work}
\label{sec:related}

\paragraph{Vision foundation models in Earth observation}
Existing pretrained models predominantly focus on RGB data, such as GFM~\cite{mendieta2023towards}, Scale-MAE~\cite{reed2023scale}, and Cross-Scale MAE~\cite{tang2024cross}. Others specialize in multispectral imagery, with FG-MAE~\cite{wang2023feature} and SatMAE~\cite{cong2022satmae} tailored for Sentinel-2 data, while SSL4EO-L~\cite{stewart2023ssl4eo} is trained on Landsat imagery. SpectralGPT~\cite{hong2024spectralgpt} leverages a 3D generative transformer for spectral data analysis. CROMA~\cite{fuller2023croma} and DeCUR~\cite{wang2024decoupling} use self-supervised multimodal learning to learn features across modalities. Satlas~\cite{bastani2023satlaspretrain} assembles a large-scale multi-sensor dataset, pretraining separate models for each sensor. DOFA~\cite{xiong2024neural} introduces a neural plasticity-inspired hypernetwork that adapts to different sensor wavelengths, enabling joint training on five distinct sensors. OmniSat~\cite{astruc2024omnisat} proposes to fuse features across multiple EO modalities to learn robust multimodal representations without labeled data. SkySense~\cite{guo2024skysense} utilizes separate visual encoders for different data modalities and designs a dedicated module for multimodal fusion.

\paragraph{Vision-language foundation models} Zero-shot foundation models like CLIP~\cite{radford2021learning} have significantly advanced computer vision. Building upon CLIP, SigLIP~\cite{zhai2023sigmoid} introduces a pairwise sigmoid loss function to enlarge the training batch and enhance the model performance. Zero-shot foundation models have also been investigated in EO applications. SkyScript~\cite{wang2024skyscript} constructs a large-scale image--text dataset by linking remote sensing images with OpenStreetMap~\cite{OpenStreetMap} semantics. RemoteCLIP~\cite{liu2024remoteclip} introduces an image--text dataset by generating captions from existing annotated datasets. GeoRSCLIP~\cite{zhang2024rs5m} introduces RS5M, an image--text dataset with 5M pairs, created by filtering existing datasets and generating captions for label-only remote sensing data. Despite these advances, existing vision--language models mainly focus on RGB data and have limitations in learning representations from multiple EO data modalities. To unify multiple data modalities, ImageBind~\cite{girdhar2023imagebind} and TaxaBind~\cite{sastry2025taxabind} align multiple modalities within a shared embedding space. LanguageBind~\cite{zhu2023languagebind} uses language as a universal intermediary for modality alignment. In the era of Large Language Models (LLMs), methods such as GeoChat~\cite{kuckreja2024geochat}, SkyEyeGPT~\cite{zhan2025skyeyegpt}, and TeoChat~\cite{irvin2024teochat} integrate CLIP-based vision models with LLMs to unify diverse tasks and achieve impressive performance.

\section{GeoLangBind-2M Dataset}
\label{sec:method}

\begin{table}[ht!]
    \centering
    \caption{Summary of dataset sources of GeoLangBind-2M. * indicates image--text datasets created in this work. SAR = synthetic aperture radar, MSI = multispectral imagery, HSI = hyperspectral imagery, and IR = infrared.}
    \resizebox{\columnwidth}{!}{%
    \begin{tabular}{l c c}
        \toprule
        \textbf{Image--Text Datasets} & \textbf{Knowledge Domain} & \textbf{\# Samples} \\
        \midrule
        Seg4 & General segmentation & 41,172 \\
        Det10 & General detection & 110,800 \\
        Flair2-RGB-caption* & Land cover analysis & 61,711 \\
        VRS-train & General detection & 20,264 \\
        SkyScript dataset & OpenStreetMap tags & 1,518,888 \\
        \midrule
        \multicolumn{3}{l}{\textbf{SAR Datasets}} \\
        MMflood-caption* & Flood mapping & 6,181 \\
        SAR-ship-caption* & Ship detection & 5,984 \\
        ChatEarthNet-SAR-caption* & Land cover analysis & 95,620 \\
        \midrule
        \multicolumn{3}{l}{\textbf{MSI/HSI Datasets}} \\
        ChatEarthNet-S2-caption* & Land cover analysis & 95,620 \\
        NLCD-hyper-caption* & Land cover analysis & 15,000 \\
        \midrule
        \multicolumn{3}{l}{\textbf{Elevation Dataset}} \\
        Flair2-Elevation-caption* & Digital Elevation Model & 61,711 \\
        \midrule
        \multicolumn{3}{l}{\textbf{Infrared Dataset}} \\
        IR-ship-caption* & Ships in infrared imagery & 18,032 \\
        \midrule
        Total & --- & 2,050,983 \\
        \bottomrule
    \end{tabular}
    }
    \label{tab:dataset_summary}
\end{table}

In response to \highlight{RQ1}, we propose GeoLangBind-2M, a large-scale image-text dataset covering a broad range of modalities including RGB, SAR, multispectral, hyperspectral, infrared, and elevation data, as listed in Table \ref{tab:dataset_summary}. Each image is paired with a high-quality text annotation, as illustrated in Fig. \ref{fig:vis_examples}. All datasets are implemented using the TorchGeo library~\cite{Stewart_TorchGeo_Deep_Learning_2022}. In the following, we introduce the composition of the dataset. 

\paragraph{RGB image--text datasets} High-resolution RGB images with textual descriptions form a significant portion of GeoLangBind-2M. We incorporate parts of the RemoteCLIP~\cite{liu2024remoteclip} dataset: Seg4 and Det10. We also include the training split of VRSBench~\cite{li2024vrsbench}, which aggregates multiple object detection datasets. SkyScript~\cite{wang2024skyscript} is a global image--text dataset for EO, from which we include the top 30\% highest quality samples. In addition to existing datasets, we construct the Flair2-RGB-caption~\cite{garioud2023flair} dataset by leveraging the Pixtral-12B~\cite{agrawal2024pixtral} model, using semantic information as prompts to generate captions for land cover features.

\paragraph{SAR image--text datasets} To enhance the dataset with SAR--text pairs, we create MMFlood-SAR-caption~\cite{montello2022mmflood}, which contains captions emphasizing water extent and flooded regions (more details in supplementary material). Additionally, the SAR-ship-caption dataset includes captions describing ships in SAR images. ChatEarthNet-SAR-caption further extends the collection by covering diverse SAR scenes from Sentinel-1, focusing on land cover types.

\paragraph{Multispectral, hyperspectral, and elevation image--text datasets} We construct ChatEarthNet-S2-caption by summarizing long, detailed descriptions from the ChatEarthNet dataset~\cite{yuan2024chatearthnet}. This dataset contains Sentinel-2 multispectral imagery covering more than 10 land cover types. Furthermore, we introduce NLCD-hyper-caption~\cite{braham2024spectralearth}, a hyperspectral dataset with 200+ EnMAP bands~\cite{guanter2015enmap}. Like Flair2-RGB-caption, we use Pixtral-12B to generate rich captions, leveraging annotated land cover maps for enhanced descriptiveness. Similarly, we also generate captions using Pixtral-12B to construct the Flair2-Elevation-caption dataset. Further details on the construction of GeoLangBind-2M are provided in the \textbf{Supplementary materials}.
\begin{figure*}[tp]
    \centering
    \includegraphics[width=0.80\linewidth]{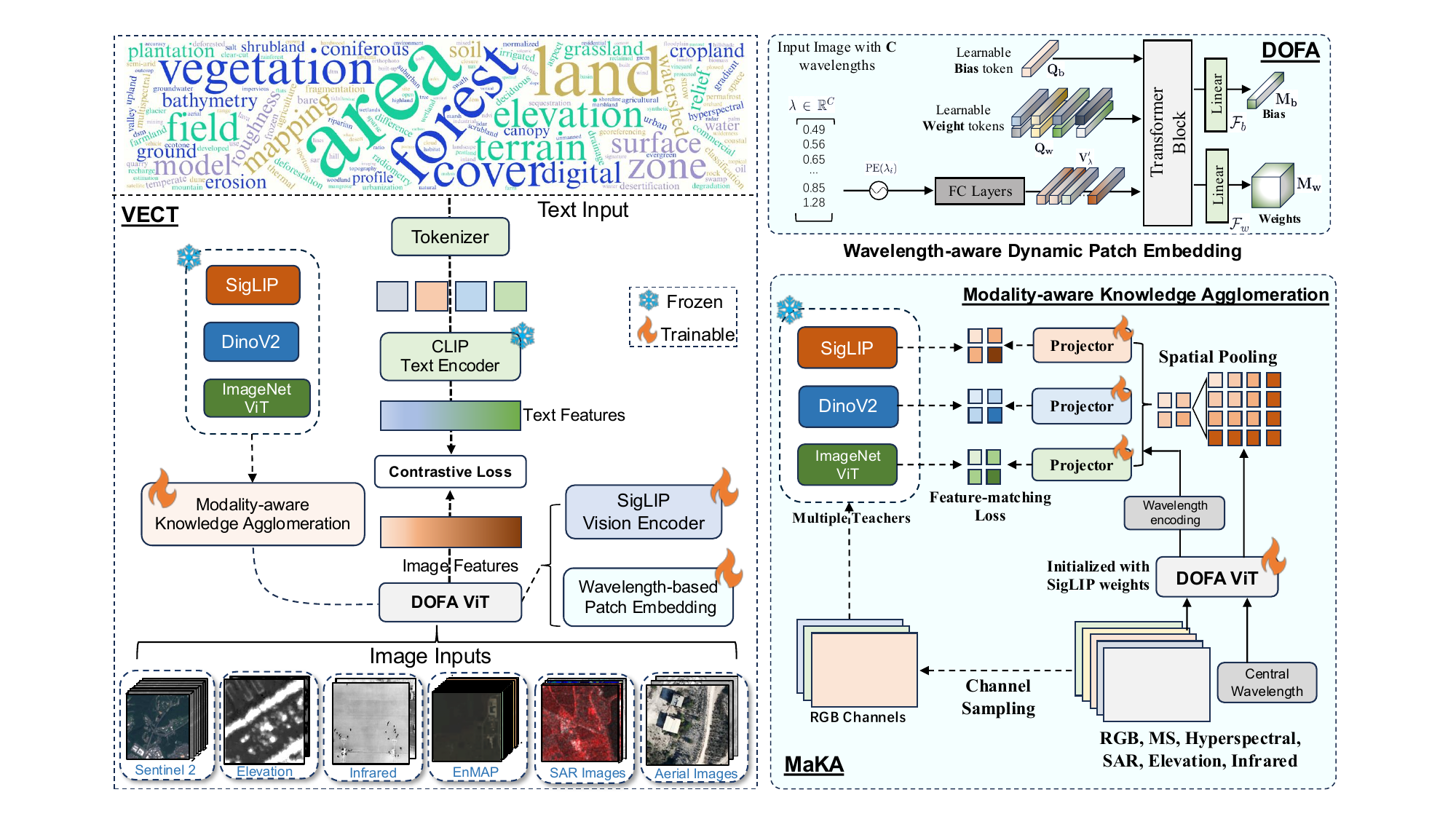}
    \caption{DOFA-CLIP aims to align diverse Earth observation data modalities into the language feature space. Vision-models Enhanced Contrastive Text-image pretraining (VECT) is proposed to enhance fine-grained image understanding. Modality-aware knowledge agglomeration (MaKA) is designed to transfer knowledge from heterogeneous sensors. The wavelength-aware dynamic patch embedding module processes multimodal images dynamically. }
    \label{fig:main_arch}
\end{figure*}

\section{Dynamic-One-For-All CLIP}
\subsection{Wavelength-aware dynamic encoder}
To handle the diversity of spectral bands across modalities, we adopt the dynamic encoder architecture from DOFA~\cite{xiong2024neural}, as illustrated in Fig. \ref{fig:main_arch}. Given an input image $\mathbf{X} \in \mathbb{R}^{C \times H \times W}$ with $C$ spectral channels, we define its corresponding central wavelengths as $\mathbf{\lambda} \in \mathbb{R}^{C}$. For modalities without well-defined wavelengths, such as elevation, we assign default wavelengths equivalent to those of the RGB channels. First, a 1D sine-cosine positional encoding is applied to embed these wavelengths into a higher-dimensional space:
$\mathbf{V}_{\lambda} = \text{PE}(\mathbf{\lambda}) \in \mathbb{R}^{C \times d_\lambda},$
where $d_\lambda$ is the embedding dimension. The encoded wavelengths $\mathbf{V}_{\lambda}$ are then transformed using two fully-connected layers with residual connections. Next, a lightweight Transformer~\cite{vaswani2017attention} encoder processes the transformed wavelength embeddings. Specifically, we concatenate the transformed embeddings $\mathbf{V'}_{\lambda}$ with the learnable weight query tokens $\mathbf{Q}_w$ and a learnable bias query token $\mathbf{Q}_b$ as input to the Transformer encoder. After the Transformer layer, the output features are passed through two fully-connected layers to generate the dynamic convolutional weights $\mathbf{M}_w$ and biases $\mathbf{M}_b$. Finally, the dynamically generated weights and biases are used for patch embedding to handle EO data modalities with varying spectral channels.

\subsection{Vision-models enhanced CLIP}
\label{vect}
To address \highlight{RQ2}, we propose VECT to enhance the spatial coherence of CLIP features by knowledge distillation from open vision models. Specifically, in addition to the contrastive loss, VECT leverages a feature-matching loss to align the intermediate features of DOFA-CLIP with features from multiple pre-trained vision models: SigLIP~\cite{zhai2023sigmoid}, DINOv2~\cite{oquab2023dinov2}, and ViT~\cite{dosovitskiy2020image}. As illustrated in Fig. \ref{fig:main_arch}, we extract three RGB channels from multispectral or hyperspectral images as input to the teacher model. The student features $\mathbf{F}_s \in \mathbb{R}^{d_t\times H'\times W'}$ are then matched to the teacher's features $\mathbf{F}_t \in \mathbb{R}^{d_t\times H'\times W'}$ using a combination of \(L_1\) loss, cosine embedding loss, and mean squared error (MSE) loss:
\begin{equation}
L_{\text{match}} = L_{L_1}(\mathbf{F}_s,\mathbf{F}_t) + L_{\text{mse}}(\mathbf{F}_s,\mathbf{F}_t) + L_{\text{cos}}(\mathbf{F}_s,\mathbf{F}_t).
\end{equation}

DOFA-CLIP is trained on GeoLangBind-2M using both the pairwise sigmoid loss $L_{\text{siglip}}$ for contrastive learning and distillation loss for feature alignment. The total loss function is:
\begin{equation}
   L = L_{\text{siglip}} + \alpha_s L_{\text{match}}^{\text{siglip}} + \alpha_d L_{\text{match}}^{\text{dinov2}} + \alpha_v L_{\text{match}}^{\text{vit}},
\end{equation}
where \(\alpha_s, \alpha_d,\) and \(\alpha_v\) are balancing factors for feature-matching losses of SigLIP, DINOv2, and ViT, respectively.

\subsection{Modality-aware knowledge agglomeration}
Considering \highlight{RQ3}, we introduce the modality-aware knowledge agglomeration (MaKA) module, which is used to translate DOFA-CLIP features in a modality-aware manner to facilitate effective feature matching in VECT. MaKA comprises three key components: a wavelength-aware prompt generator, a modality-aware conditional layer normalization, and a feature projector.

\paragraph{1. Wavelength-aware prompt generator}  
Each modality is associated with a set of wavelengths, which we encode using a sine-cosine positional encoding function. Given input wavelengths $\mathbf{\lambda} \in \mathbb{R}^{C}$, we first obtain their positional embeddings $\mathbf{V}_{\lambda} \in \mathbb{R}^{C \times d_\lambda}$.  
To generate the modality-aware prompt vector, we apply a linear projection $\mathbf{W}_{\text{proj}} \in \mathbb{R}^{d \times d_\lambda}$ to each row of $\mathbf{V}_{\lambda}$, then compute the mean over the $C$ wavelengths:
\begin{equation}
\mathbf{V}_p = \frac{1}{C} \sum_{i=1}^{C} \bigl(\mathbf{W}_{\text{proj}} \, \mathbf{V}_{\lambda_i} \bigr),
\end{equation}
where $\mathbf{V}_p \in \mathbb{R}^{d}$ is the fused modality embedding. This final embedding vector $\mathbf{V}_p$ captures modality-specific information derived from the provided wavelengths.

\paragraph{2. Modality-aware layer normalization}
Given the feature map from the last layer of the DOFA-CLIP model, we first interpolate it to match the teacher's feature resolution, denoted as \(\mathbf{F} \in \mathbb{R}^{d \times H' \times W'}\), omitting the batch dimension for simplicity. With $\mathbf{F}$ and a modality prompt \(\mathbf{V}_p \in \mathbb{R}^{d}\), our goal is to perform layer normalization that dynamically adapts to the modality. To achieve this, we introduce modality-aware conditional layer normalization as follows.

First, we derive two learnable modulation vectors,  
\(\boldsymbol{\gamma}, \boldsymbol{\beta} \in \mathbb{R}^{d}\),  
from the modality prompt via a linear transformation:
\begin{equation}
\begin{bmatrix}
\boldsymbol{\gamma} \\[6pt]
\boldsymbol{\beta}
\end{bmatrix}
\;=\;
\mathbf{W}_{\text{prompt}} \, \mathbf{V}_p
\;\in\;\mathbb{R}^{2d},
\end{equation}
where \(\mathbf{W}_{\text{prompt}} \in \mathbb{R}^{2d \times d}\). We then split the output into  
\(\boldsymbol{\gamma}, \boldsymbol{\beta} \in \mathbb{R}^{d}\)  
and reshape them to broadcast over the spatial dimensions:
\begin{equation}
\boldsymbol{\gamma} \;\rightarrow\; (1,\, d,\, 1,\, 1),
\quad
\boldsymbol{\beta} \;\rightarrow\; (1,\, d,\, 1,\, 1).
\end{equation}

Let \(\mathrm{LN}(\mathbf{F})\) denote the standard channel-wise layer normalization of \(\mathbf{F}\). The modality-aware layer normalization is then defined as:
\begin{equation}
\mathbf{F}' \;=\; \mathrm{LN}(\mathbf{F}) + \boldsymbol{\gamma} \odot \mathrm{LN}(\mathbf{F}) \;+\; \boldsymbol{\beta},
\end{equation}
where \(\odot\) denotes the Hadamard product. Note that we use the modality-aware layer normalization to learn the residual information specific to each modality.

\paragraph{3. Projector module}  
A convolutional layer is applied to adjust the feature representation \(\mathbf{F}'\) to channel dimensions $d_t$ of the corresponding teachers. The final representation is then used to compute the feature-matching loss for model distillation. This design ensures that the learned features are both \emph{aligned} and \emph{modality-specific}, enhancing robust adaptation across different input modalities.

To address the data imbalance issue, we employ a \textbf{weight space merging} strategy~\cite{ilharco2022patching}. Since RGB data is the most dominant modality, we split the dataset into two parts: RGB data and non-RGB data. We first train two separate DOFA-CLIP models on these subsets. Then, we merge their weights to obtain a unified model that can handle both RGB and other modalities. Since the original SigLIP model is trained on massive data, we also merge the original SigLIP weights with our model. 
For clarity, let
\(\theta_{\text{siglip}}\) be the original SigLIP model weights,
\(\theta_{\text{rgb}}\) be the DOFA-CLIP model weights trained on the RGB subset,
\(\theta_{\text{others}}\) be the DOFA-CLIP model weights trained on the other five modalities. In the first step, we merge \(\theta_{\text{siglip}}\) and \(\theta_{\text{rgb}}\) using a simple linear weight merging strategy:
\begin{equation}
\theta^{*} = (1 - m_1) \, \theta_{\text{siglip}} + m_1 \, \theta_{\text{rgb}},
\end{equation}
where \(m_1\) controls the weighting ratio between the two models. Next, we merge the intermediate weights \(\theta^{*}\) with \(\theta_{\text{others}}\) to obtain the final model:
\begin{equation}
\theta = (1 - m_2) \, \theta^{*} + m_2 \, \theta_{\text{others}},
\end{equation}
where \(m_2\) determines the contribution of the non-RGB modalities. We conduct ablation studies by varying \(m_1\) and \(m_2\) to search for optimal weighting ratios. While more advanced merging methods exist~\cite{akiba2025evolutionary}, we use a linear approach to assess its effectiveness on heterogeneous EO data, leaving sophisticated strategies for future work.

\begin{table*}[ht]
    \centering
    \caption{Zero-shot comparison of various models on scene and fine-grained classification tasks. Bold values indicate the best performance.}
    \small
    \renewcommand{\arraystretch}{1.0}
    \scalebox{0.7}
    {
    \begin{tabular}{cccccccccccccc}
        \toprule
        \multirow{2}{*}{ViT} & \multirow{2}{*}{Model} & \multicolumn{9}{c}{Scene classification} & \multicolumn{3}{c}{Fine-grained classification} \\
        \cmidrule(lr){3-11} \cmidrule(lr){12-14}
        & & SkyScript & AID & EuroSAT & fMoW & Million-AID & PatternNet & RESISC & RSI-CB & Avg. & Roof shape & Smoothness & Surface \\
        \midrule
        \multirow{6}{*}{Base} 
        & CLIP-original & 40.16 & 69.55 & 32.11 & 17.62 & 57.27 & 64.09 & 65.71 & 41.26 & 49.66 & 31.50 & 26.80 & 61.36 \\
        & Human-curated captions & 40.03 & 71.05 & 33.85 & 18.02 & 57.48 & 66.56 & 66.04 & 42.73 & 50.82 & 28.50 & 27.80 & 60.91 \\
        & RemoteCLIP & 27.06 & \textbf{87.05} & 30.74 & 11.13 & 46.26 & 56.05 & 67.88 & 44.55 & 49.09 & 30.50 & 21.00 & 43.86 \\
        & CLIP-laion-RS & 40.77 & 69.55 & 37.63 & 19.16 & 56.59 & 64.79 & 64.63 & 41.79 & 50.59 & 28.83 & 27.60 & 62.27 \\
        & SkyCLIP-50 & 52.98 & 70.90 & 33.30 & 19.24 & 62.69 & 72.18 & 66.67 & 46.20 & 53.02 & 26.00 & \textbf{38.00} & {67.73} \\
        \midrule
        & DOFA-CLIP-B-224 & \textbf{70.39} & {77.60} & \textbf{52.30} & \textbf{20.17} & \textbf{64.91} & \textbf{76.68} & \textbf{67.21} & \textbf{49.53} & \textbf{59.85} & \textbf{44.33} & 20.00 & \textbf{72.73} \\
        \midrule
        \multirow{8}{*}{Large} 
        & CLIP-original & 55.06 & 69.25 & 41.89 & 26.19 & 57.88 & 71.39 & 66.70 & 43.02 & 53.76 & 37.50 & 25.40 & 42.73 \\
        & Human-curated captions & 56.09 & 72.95 & 41.96 & 26.33 & 58.47 & 74.86 & 68.70 & 44.60 & 55.41 & 37.00 & 26.60 & 40.00 \\
        & RemoteCLIP & 34.40 & 70.85 & 27.81 & 16.77 & 47.20 & 61.91 & \textbf{74.31} & 50.79 & 49.99 & 34.33 & 34.20 & 55.45 \\
        & CLIP-laion-RS & 58.81 & 71.70 & 54.30 & 27.21 & 60.77 & 72.68 & 71.21 & 48.21 & 57.82 & 40.50 & \textbf{37.60} & 53.41 \\
        & SkyCLIP-20 & 67.94 & 71.95 & 53.63 & 28.04 & 65.68 & 78.62 & 70.70 & 50.03 & 59.98 & 44.83 & 26.80 & 61.36 \\
        & SkyCLIP-30 & 69.08 & 72.15 & 52.44 & 27.77 & 66.40 & 79.67 & 70.77 & 50.19 & 59.99 & 46.17 & 30.80 & 64.32 \\
        & SkyCLIP-50 & 70.89 & 71.70 & 51.33 & 27.12 & 67.45 & \textbf{80.88} & 70.94 & 50.09 & 59.93 & 46.83 & 35.80 & 67.50 \\
        \midrule
        & DOFA-CLIP-L-384 & \textbf{76.83} & \textbf{75.50} & \textbf{59.04} & \textbf{29.10} & \textbf{70.16} & {80.17} & {73.15} & \textbf{51.62} & \textbf{64.45} & \textbf{61.83} & 26.00 & \textbf{81.36} \\
        \bottomrule
    \end{tabular}
    }
    \label{tab:zeroshot1}
\end{table*}

\begin{table*}[ht]
    \centering
    \caption{Zero-shot comparison on GeoBench datasets with non-RGB data. Bold values indicate the best performance.}
    \small
    \renewcommand{\arraystretch}{1.0}
    \scalebox{0.8}
    {
    \begin{tabular}{ccccccccc}
        \toprule
        \multirow{2}{*}{Model (Base version)} & \multicolumn{3}{c}{m-bigearthnet} & \multicolumn{2}{c}{m-so2sat} & \multicolumn{2}{c}{m-forestnet} \\
        \cmidrule(lr){2-4}  \cmidrule(lr){5-6} \cmidrule(lr){7-8}
        & Precision & Recall & F1 & RGB & Sentinel-2 (5 bands) & RGB & Landsat-8 (5 bands) \\
        \midrule
        SigLIP & 45.34 & 12.36 & 16.82 & 12.88 & - & 8.16 & - \\
        RemoteCLIP & 33.82 & 20.35 & 18.84 & 10.75 & - & 8.46 & - \\
        SkyCLIP-50 & 39.91 & 19.58 & 20.32 & 11.97 & - & 10.78 & - \\
        \midrule
        DOFA-CLIP-B (224) & \textbf{47.70} & \textbf{20.37} & \textbf{23.69} & \textbf{17.95} & 14.60 & \textbf{13.60} & \textbf{17.02}\\
        \bottomrule
    \end{tabular}
    }
    \label{tab:zeroshot2}
\end{table*}

\section{Experiments}

\subsection{Implementation details}
DOFA-CLIP is trained on 8 NVIDIA A100 40GB GPUs for the base model and 8 NVIDIA A100 80GB GPUs for the large version. For the base version, we use a ViT-Base architecture (patch size 16) pretrained on WebLI~\cite{chen2022pali}. The teacher models are ViT-Base-SigLIP (224), DINOv2-ViT-L/14, and ViT-Huge (ImageNet). The large version is based on SoViT-400m~\cite{alabdulmohsin2023getting} (patch size 14, input size 384), with teacher models ViT-So400m-SigLIP (384), DINOv2-ViT-L/14, and ViT-Large (ImageNet). Training is conducted for 20 epochs using AdamW (learning rate 5e-4, weight decay 1e-7).

\subsection{Performance comparison}
We follow the evaluation protocols of \citet{wang2024skyscript} and evaluate the models via zero-shot classification tasks. Specifically, the following eleven image scene classification datasets are used: AID~\cite{xia2017aid}, EuroSAT~\cite{helber2019eurosat}, fMoW~\cite{christie2018functional}, Million-AID~\cite{long2021creating}, PatternNet~\cite{zhou2018patternnet}, NWPU-RESISC45~\cite{cheng2017remote}, RSI-CB256~\cite{li2017rsi}, as well as the fine-grained attribute classification datasets. In addition, we further compare these models on some datasets from the GEO-Bench suite~\cite{lacoste2024geo}: m-bigearthnet~\cite{sumbul2019bigearthnet}, m-so2sat~\cite{zhu2019so2sat}, and m-forestnet~\cite{irvin2020forestnet}.

\begin{table*}[t]
\centering
\caption{Recall@1, recall@5, and recall@10 for cross-modal retrieval on three benchmark datasets. Bold values indicate the best performance.}
\resizebox{\textwidth}{!}{%
\begin{tabular}{lccccccccccccccccccc}
\toprule
& \multicolumn{6}{c}{\textbf{RSICD}} 
& \multicolumn{6}{c}{\textbf{RSITMD}} 
& \multicolumn{6}{c}{\textbf{UCM-caption}} \\
\cmidrule(lr){2-7}\cmidrule(lr){8-13}\cmidrule(lr){14-19}
& \multicolumn{3}{c}{Image to Text} 
& \multicolumn{3}{c}{Text to Image} 
& \multicolumn{3}{c}{Image to Text} 
& \multicolumn{3}{c}{Text to Image} 
& \multicolumn{3}{c}{Image to Text} 
& \multicolumn{3}{c}{Text to Image} \\
\cmidrule(lr){2-4}\cmidrule(lr){5-7}\cmidrule(lr){8-10}\cmidrule(lr){11-13}\cmidrule(lr){14-16}\cmidrule(lr){17-19}
\textbf{Model} 
& R@1 & R@5 & R@10 & R@1 & R@5 & R@10
& R@1 & R@5 & R@10 & R@1 & R@5 & R@10
& R@1 & R@5 & R@10 & R@1 & R@5 & R@10 \\
\midrule
\multicolumn{19}{l}{\emph{RSICD, RSITMD, and/or UCM-caption seen in training}} \\
\midrule
AMFMN~\cite{yuan2022exploring}
& 5.39 & 15.08 & 23.40 & 4.90 & 18.28 & 31.44 
& 10.63 & 24.78 & 41.81 & 11.51 & 34.69 & 54.87 
& 16.67 & 45.71 & 68.57 & 12.86 & 53.24 & 79.43 \\

LW-MCR-u~\cite{yuan2021lightweight}
& 4.39 & 13.35 & 20.29 & 4.30 & 18.85 & 32.34 
& 9.73 & 26.77 & 37.61 & 9.25 & 34.07 & 54.03 
& 18.10 & 47.14 & 63.81 & 13.14 & 50.38 & 79.52 \\

GaLR~\cite{yuan2022remote}
& 6.59 & 19.85 & 31.04 & 4.69 & 19.48 & 32.13 
& 14.82 & 31.64 & 42.48 & 11.15 & 36.68 & 51.68 
& --   & --    & --    & --   & --    & --    \\

\midrule
\multicolumn{19}{l}{\emph{RSICD, RSITMD, and/or UCM-caption not seen in training}} \\
\midrule
CLIP-original 
& 6.59 & 20.68 & 31.75 & 3.62 & 14.28 & 23.63
& 10.18 & 30.31 & 42.04 & 8.31 & 24.96 & 39.03
& 37.62 & 78.10 & 89.52 & 28.12 & 64.99 & 77.19 \\

CLIP-laion-RS
& 8.42 & 23.70 & 35.86 & 5.81 & 19.49 & 30.25
& 13.72 & 32.08 & 44.91 & 10.57 & 31.48 & 46.96
& 39.52 & 79.52 & 90.00 & 29.71 & 62.60 & 80.37 \\

SkyCLIP-30  
& \textbf{8.97} & 24.15 & \textbf{37.97} & 5.85 & 20.53 & 33.53
& 11.73 & 33.19 & 47.35 & 10.19 & 32.47 & 49.08
& 38.57 & \textbf{84.29} & 93.81 & 31.83 & 64.19 & 81.96 \\

\midrule
DOFA-CLIP-L
& 8.42 & \textbf{25.16} & 37.05 & \textbf{8.15} & \textbf{24.90} & \textbf{37.73}
& \textbf{13.94} & 30.31 & 44.47 & \textbf{13.45} & \textbf{38.70} & \textbf{55.78}
& \textbf{43.33} & 81.43 & \textbf{93.81} & \textbf{35.28} & \textbf{71.62} & \textbf{87.80} \\
\bottomrule
\end{tabular}
}
\label{tab:retreival}
\end{table*}

\paragraph{Overall zero-shot performance} As shown in Table~\ref{tab:zeroshot1}, DOFA-CLIP establishes new state-of-the-art results in zero-shot remote-sensing image classification. Compared with existing CLIP-based baselines such as CLIP-original, RemoteCLIP, CLIP-laion-RS, and the various SkyCLIP configurations, DOFA-CLIP-B-224 consistently outperforms all counterparts. Its average accuracy across the eight scene datasets is significantly higher than that of the other base models. The Larger variant of DOFA-CLIP with higher resolution yields further accuracy gains. On three fine-grained roof-attribute classification tasks, both DOFA-CLIP variants achieve strong accuracy on roof shape and surface, significantly outperforming other methods. Although the results on smoothness remain more modest, our approach still provides a clear improvement over existing approaches.

\paragraph{Zero-shot classification on multispectral imagery} The zero-shot classification results on three GEO-Bench datasets are presented in Table \ref{tab:zeroshot2}. DOFA-CLIP demonstrates competitive or superior zero-shot multi-label classification performance compared to existing models on the m-bigearthnet dataset (43 classes). For the m-so2sat dataset (17 classes), when utilizing five spectral bands, the performance decreases slightly due to the significantly less multispectral training data compared to RGB. \emph{Note that competitive models cannot process 5-band images as input for zero-shot classification}. On the m-forestnet dataset (12 classes), DOFA-CLIP substantially outperforms other models and shows remarkable accuracy improvements when using five spectral bands as input. Notably, the m-forestnet dataset uses Landsat-8 data, which is unseen in our training dataset.

\paragraph{Cross-modal retrieval performance} 
Table \ref{tab:retreival} compares DOFA-CLIP against recent methods on three benchmark caption datasets: RSICD, RSITMD, and UCM-caption. We report Recall@1, Recall@5, and Recall@10 for both image-to-text and text-to-image retrieval. Despite the inherent difficulty of remote-sensing cross-modal tasks, DOFA-CLIP-L consistently achieves competitive or higher recall scores than most competitive baselines, including CLIP-original, CLIP-laion-RS, and SkyCLIP-30.

\begin{table}[ht]
    \centering
    \caption{Ablation experiments of models on the AID zero-shot classification dataset.}
    \scalebox{0.68}{
    \begin{tabular}{lccc}
    \toprule
    Model &  Distill. & Top-1 Accuracy & Top-5 Accuracy \\
    \midrule
    SigLIP & -- & 65.95 & 96.20 \\                  
    \midrule
    DOFA-CLIP-B (RGB) &\cmark & 76.90 & 94.55 \\            
    SigLIP+DOFA-CLIP-B (RGB) &\cmark & \textbf{77.70} & 95.50 \\   
    SigLIP+DOFA-CLIP-B (RGB) &\xmark & 72.00 & 95.80 \\ 
    \midrule
    DOFA-CLIP-B (Others) & \cmark & 64.70 & 91.95 \\         
    SigLIP+DOFA-CLIP-B (Others) & \cmark & 69.50 & \underline{96.20} \\    
    SigLIP+DOFA-CLIP-B (Others) & \xmark & 66.30 & 91.20 \\ 
    \midrule
    DOFA-CLIP-B (Mixed) (baseline) &\cmark & 71.60 & 93.65 \\
    DOFA-CLIP-B (Merging) (ours) &\cmark & \underline{77.60} & \textbf{96.65} \\
    \bottomrule
    \end{tabular}
    }
    \label{tab:ablation1}
\end{table}
\subsection{Ablation studies}
\label{ablation}
\begin{table}[ht]
    \centering
    \caption{Ablation experiments of different distillation methods on the segmentation datasets.}
    \scalebox{0.78}{
    \begin{tabular}{lccc}
    \toprule
    Model &  MADOS & m-nz-cattle & m-NeonTree \\
    \midrule
    SigLIP & 57.2 & 77.5 & 52.3 \\                  
    \midrule
    DOFA-CLIP-B w/o MaKA &61.7 & 81.5 & 58.1 \\            
    DOFA-CLIP-B w/ MaKA &\textbf{62.3} & \textbf{82.2} & \textbf{59.0} \\   
    \bottomrule
    \end{tabular}
    }
    \label{tab:ablation2}
\end{table}

\begin{table}[ht]
    \centering
    \caption{Ratio for merging SigLIP and DOFA-CLIP-B (RGB).}
    \scalebox{0.8}
    {
    \begin{tabular}{cccc}
    \toprule
    $m_1$ & AID (Top-1) & RSI-CB (Top-1) & Roof shape (Top-1)  \\
    \midrule
    0 & 65.95 & 38.01 & 41.17  \\
    0.1 & 44.25 & 27.33 & 37.67  \\
    0.3 & 56.75  & 31.87 & 38.00  \\
    0.5 & 74.65 & 41.80 & 44.33  \\
    0.7 & \underline{77.40} & 45.36 & 53.67  \\
    \rowcolor{gray!25} 0.9 & \textbf{77.70} & \textbf{47.03} & \underline{57.83}  \\
    1.0 & 76.90 & \underline{46.72} & \textbf{58.50}  \\
    \bottomrule
    \end{tabular}
    }
    \label{tab:ratios1}
\end{table}

\paragraph{Weight merging and MaKA}
Table \ref{tab:ablation1} presents the results of our ablation experiments, evaluating the effects of weight merging and MaKA across different models on the AID zero-shot classification dataset. The baseline SigLIP model can be significantly improved by incorporating the DOFA-CLIP-B (RGB) model, trained on the RGB part of GeoLangBind-2M. The model trained with our MaKA module (for distillation) can largely outperform the non-distilled version. DOFA-CLIP-B (Others), trained on non-RGB modalities, performs lower than its RGB counterpart. When merged with SigLIP, the performance can be significantly improved. DOFA-CLIP-B (Mixed) is the model trained with all the data modalities mixed together. When compared with the DOFA-CLIP-B (Merging), the performance is significantly lower. This demonstrates that weight merging is more effective and flexible than direct mixing of data modalities.
\begin{figure}[tp]
    \centering
    \includegraphics[width=0.98\linewidth]{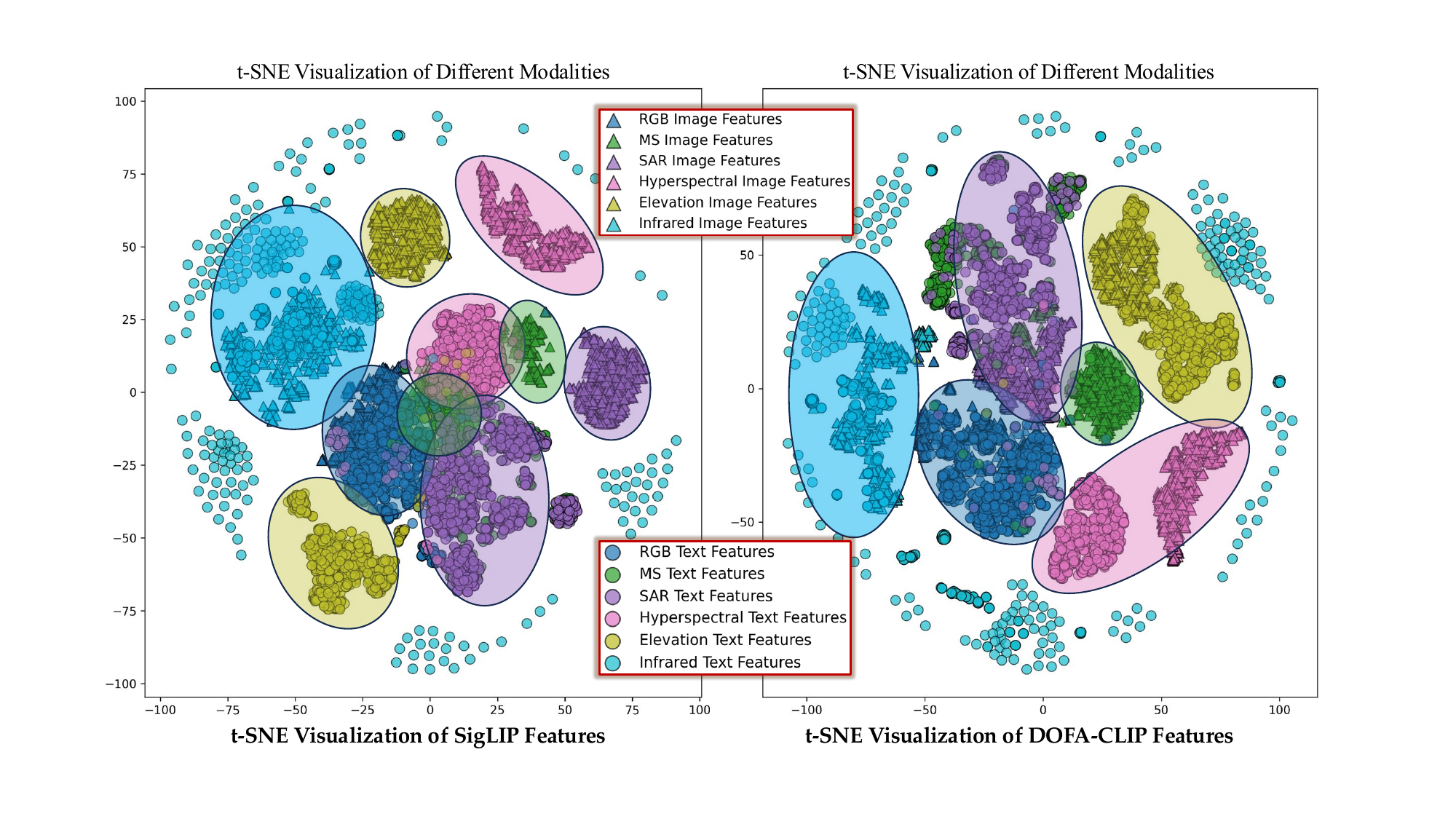}
    \caption{t-SNE visualization of feature distributions for SigLIP features (left) and DOFA-CLIP-L-384 model (right).}
    \label{fig:tsne}
\end{figure}
\paragraph{MaKA enhances segmentation performance}
As presented in Table \ref{tab:ablation2}, we conducted ablation studies on the MADOS~\cite{kikaki2024detecting}, m-nz-cattle~\cite{abuaiadah2022remote}, and m-NeonTree~\cite{weinstein2021benchmark} datasets. DOFA-CLIP-B without MaKA represents a model trained using a baseline distillation method that does not incorporate modality-specific wavelengths as conditioning. The results demonstrate the significant effectiveness of the MaKA module, showing consistent performance improvements across all datasets. \textbf{More details are in the supplementary material.}

\paragraph{Weight merging ratios} In Table \ref{tab:ratios1} and \ref{tab:ratios2}, a linear search is conducted to determine the optimal ratios for weight merging. Based on the results, we select 0.9 for $m_1$ to merge the weights of SigLIP and DOFA-CLIP (RGB). We choose 0.5 for merging weights of the intermediate weights and DOFA-CLIP (Others). Thus, we set $m_1=0.9$ and $m_2=0.5$ to derive our final model weights. \emph{Note that the weight merging is executed a single time using a designated zero-shot classification dataset, and held fixed across all tasks.}

\begin{table}[ht]
    \centering
    \caption{Ratio for merging weights of DOFA-CLIP-B (RGB) and DOFA-CLIP-B (Others).}
    \scalebox{0.8}
    {
    \begin{tabular}{cccc}
    \toprule
    $m_2$ & AID (Top-1) & RSI-CB (Top-1) & Roof shape (Top-1)  \\
    \midrule
    0 & 77.70 & 47.03 & 57.83  \\
    0.1 & 78.10 & 47.31 & 57.50  \\
    0.3 & \underline{78.55} & 47.91 & 55.83  \\
    \rowcolor{gray!25} 0.5 & \textbf{78.60} & \underline{49.17} & 52.67  \\
    0.7 & 76.45 & \textbf{49.92} & 52.00  \\
    0.9 & 76.35 & 43.98 & \textbf{63.33}  \\
    1.0 & 69.20 & 34.80 & \underline{59.67}  \\
    \bottomrule
    \end{tabular}
    }
    \label{tab:ratios2}
\end{table}

\begin{figure}[htbp!]
    \centering
    \includegraphics[width=0.98\linewidth]{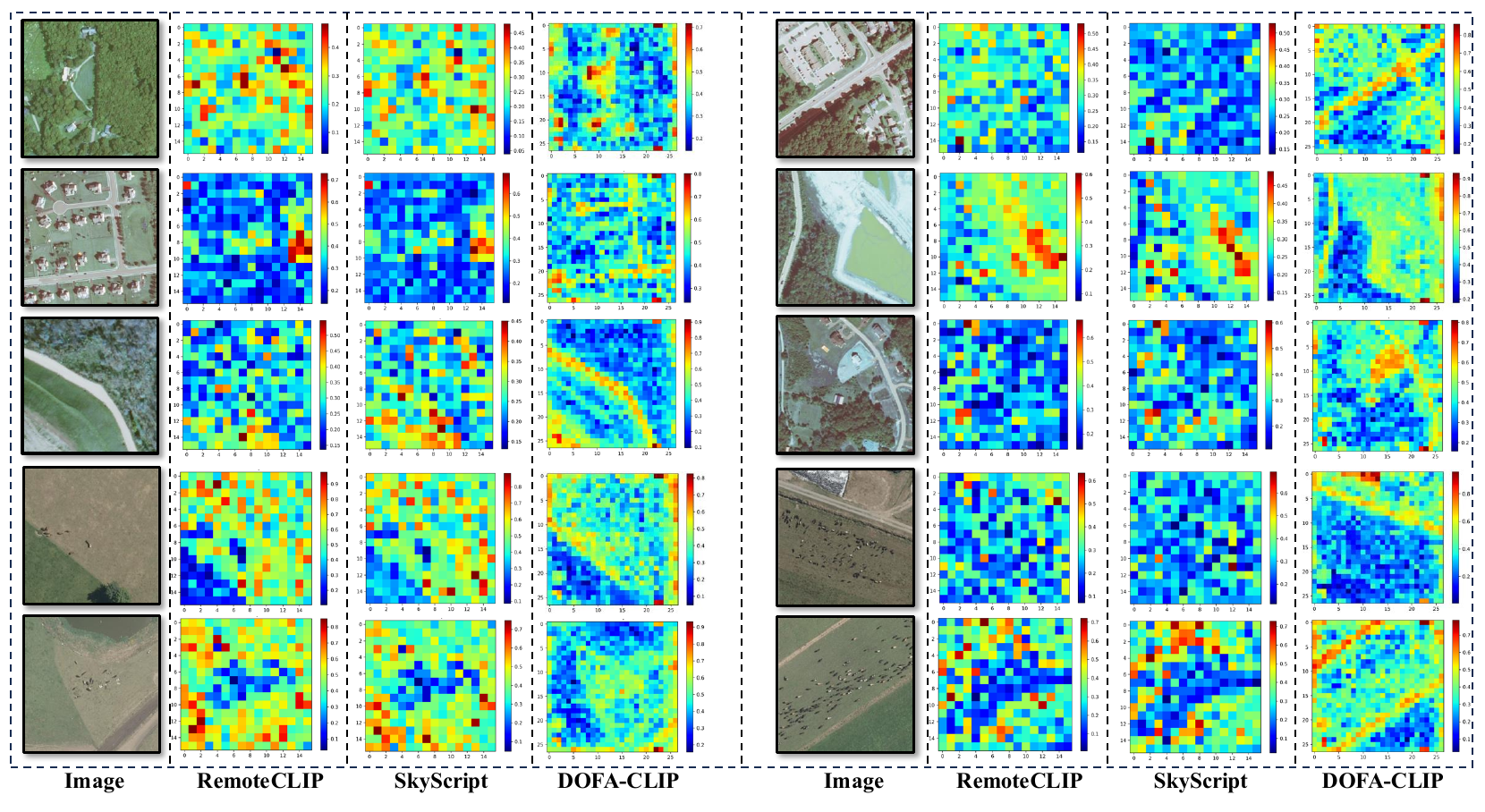}
    \caption{Visual comparison of deep features from RemoteCLIP, Skyscript, and DOFA-CLIP.}
    \label{fig:vis_feat}
\end{figure}
\subsection{Feature visualization and analysis}
On the left side of Fig. \ref{fig:tsne}, we present the t-SNE plot of SigLIP features. Compared to the SigLIP features, the features of DOFA-CLIP-L-384 demonstrate a significantly stronger imagery-text alignment across all modalities, making it a powerful foundation for Earth observation tasks. In Fig. \ref{fig:vis_feat}, we visually compare the feature heatmaps of RemoteCLIP, SkyScript, and DOFA-CLIP. DOFA-CLIP features are more spatially structured and preserve more object details, demonstrating the effectiveness of VECT.

\section{Conclusion}
\label{sec:conclusion}
We introduce DOFA-CLIP, a unified vision--language foundation model that bridges heterogeneous EO modalities through a shared language space. To achieve this, we construct GeoLangBind-2M, a large-scale image--text dataset containing six EO data modalities. DOFA-CLIP consists of a wavelength-aware encoder and a modality-aware knowledge agglomeration module to enhance fine-grained image understanding. Additionally, progressive weight merging is proposed to scale the model training to multiple data modalities. Extensive experiments demonstrate the state-of-the-art performance of DOFA-CLIP across zero-shot classification, semantic segmentation, and retrieval tasks.

\section{Supplementary material}
\label{sec:appendix_section}

The supplementary material contains the following sections:
\begin{enumerate}
    \item Details of the evaluation datasets;
    \item Details on the construction of GeoLangBind-2M;
    \item More visualization examples of  GeoLangBind-2M;
    \item Ablation studies on the distillation loss terms;
    \item Confusion matrices of the zero-shot classification tasks;
    \item More visualization of the features;
\end{enumerate}

\subsection{Details of the evaluation datasets}
We evaluate DOFA-CLIP on both datasets that have been established in previous works for direct comparison, as well as additional datasets that highlight the wide-ranging capabilities of the proposed model. 

The following datasets are part of the SkyScript~\cite{wang2024skyscript} evaluation framework, which we follow for direct comparison.
\begin{itemize}
    \item SkyScript classification dataset~\cite{wang2024skyscript}: contains 7,000 aerial RGB images with 70 classes and different objects than the SkyScript pretraining dataset.
    \item AID~\cite{xia2017aid}: contains 2,000 aerial RGB images and 30 image scene classes. 
    \item EuroSAT~\cite{helber2019eurosat}: contains 27,000 Sentinel-2 images with 13 spectral bands and 10 image scene classes. We use the dedicated test split that contains 2,700 images.
    \item fMoW~\cite{christie2018functional}: contains 106,081 RGB aerial images with 62 image scene classes.
    \item Million-AID~\cite{long2021creating}: contains 10,000 RGB aerial images of varying GSD with 51 image scene classes.
    \item PatternNet~\cite{zhou2018patternnet}: contains 30,400 high-resolution RGB aerial images with 6--50~cm GSD and 38 image scene classes.
    \item NWPU-RESISC45~\cite{cheng2017remote}: contains 31,500 aerial RGB images with 0.2--30~m GSD and 45 image scene classes.
    \item RSI-CB~\cite{li2017rsi}: contains 24,747 aerial RGB images and 35 image scene classes.
\end{itemize}

\begin{table*}[h!]
    \centering
    \scriptsize
    \caption{GEO-Bench dataset description.}
    \begin{tabular}{lc|ccccccc}
        \toprule
        Tasks & \textbf{Name} & \textbf{Image Size} & \textbf{\# Classes} & \textbf{Train} & \textbf{Val} & \textbf{Test} & \textbf{\# Bands} & \textbf{Sensors} \\
        \midrule
        \multirow{3}{*}{\rotatebox{0}{Classification}} & m-bigearthnet & 120 x 120 & 43 & 20000 & 1000 & 1000 & 12 & Sentinel-2 \\
        & m-so2sat & 32 x 32 & 17 & 19992 & 986 & 986 & 18 & Sentinel-2 + Sentinel-1 \\
        & m-forestnet & 332 x 332 & 12 & 6464 & 989 & 993 & 6 & Landsat-8 \\
        \midrule
        \multirow{2}{*}{\rotatebox{0}{Segmentation}} 
        & m-nz-cattle & 500 x 500 & 2 & 524 & 66 & 65 & 3 & RGB \\
        & m-NeonTree & 400 x 400 & 2 & 270 & 94 & 93 & 5 & RGB + Hyperspectral + Elevation \\
        \bottomrule
    \end{tabular}
    \label{tab:geobench_ds_summary}
\end{table*}

Additionally, we choose the GEO-Bench~\cite{lacoste2024geo} suite of datasets that cover a range of relevant remote sensing tasks across different domains, sensors, and geospatial locations. Table \ref{tab:geobench_ds_summary} contains a detailed overview of the number of samples, sensors, and target classes in the GEO-Bench datasets.
\begin{figure}[tp]
    \centering
    \includegraphics[width=0.85\linewidth]{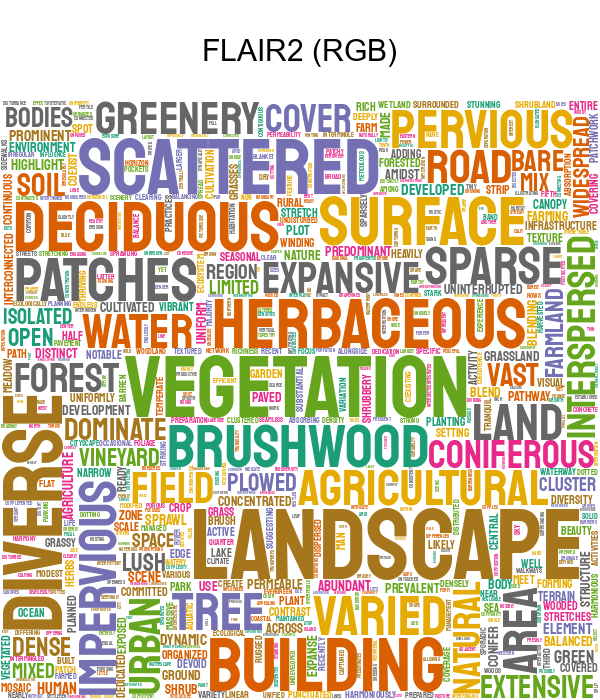}
    \caption{Word cloud for the Flair2-RGB-caption dataset.}
    \label{fig:wc_flair_rgb}
\end{figure}

\begin{figure}[tp]
    \centering
    \includegraphics[width=0.92\linewidth]{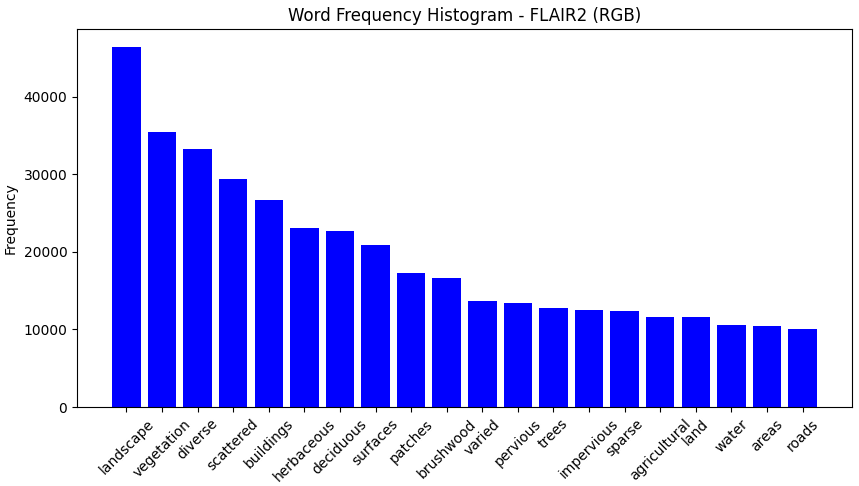}
    \caption{Word histogram for the Flair2-RGB-caption dataset.}
    \label{fig:wh_flair_rgb}
\end{figure}

In addition, we conduct ablation experiments on the Marine Debris and Oil Spill (MADOS)~\cite{kikaki2024detecting} dataset. The MADOS dataset is a high-resolution multispectral Sentinel-2 dataset for marine pollution detection. Spanning 174 globally distributed scenes (2015--2022) with 1.5M annotated pixels, it covers diverse pollutants, sea surface features, and water-related classes. Unlike existing datasets, MADOS enables scalable, generalizable deep learning models for holistic marine pollution monitoring.

\subsection{Details on constructing GeoLangBind-2M}
We construct our multimodal image--text dataset GeoLangBind-2M by assembling existing datasets and curating 8 new datasets. Details on the newly curated sub-datasets are as follows.
\subsubsection{Flair2-RGB-caption}
For the Flair2-RGB-caption dataset, originally designed for semantic segmentation land cover classification, we employed an approach similar to the ChatEarthNet~\cite{yuan2024chatearthnet} dataset construction. We select the training set of 61,711 images from FLAIR~\#2~\cite{garioud2023flair} and derive contextual information from their corresponding pixel-level segmentation masks. This extracted semantic content, paired with the segmentation maps as visual context, serves as the input for generating comprehensive descriptions using the Pixtral~12B~\cite{agrawal2024pixtral} language model.

Our semantic extraction process involves calculating the distribution percentages of various terrain categories within each image and formulating these statistics into structured contextual cues. We identify all land cover categories in each segmentation map, assigning distinctive color codes to represent different semantic classes. The designed prompt explicitly defines the correspondence between color codes and land cover categories, while also incorporating the calculated percentage coverage of each land/object type within the segmentation maps. This quantitative spatial information is then processed by the Pixtral~12B model to analyze distribution patterns and generate linguistically accurate descriptions.

With this semantic framework established, we craft a prompt for generating detailed image descriptions. The prompt template variables indicated in brackets represent dynamic content populated during the semantic processing phase. The exact prompt template used for Flair2-RGB-caption creation is shown below:

\begin{mdframed}[backgroundcolor=gray!5, linecolor=black]
You are an AI visual assistant capable of describing a scene based on a segmentation map. The map represents different land cover types using specific colors. The legend is as follows:

- [$color\_i$] corresponds to [$label\_i$], occupying [$percent\_i$]\% of the image.

Do not mention colors, color coding, or technical details. Use the given land cover class names exclusively.
Generate a brief yet natural description of the scene by extending the sentence:  
``The aerial image contains [$presented\_labels$] land types.'' 
Provide a concise summary of their spatial distribution.
\end{mdframed}

In Fig. \ref{fig:wc_flair_rgb}, we illustrate the word cloud for this dataset.

\subsubsection{Flair2-Elevation-caption}
The \textbf{Flair2-Elevation-Caption} dataset is constructed to generate detailed textual descriptions of terrain characteristics by leveraging both semantic segmentation maps and elevation maps. The dataset creation follows a structured approach, similar to the Flair2-RGB-caption dataset, ensuring that the generated captions effectively describe the topographical and land cover features of each image.
\begin{figure}[tp]
    \centering
    \includegraphics[width=0.85\linewidth]{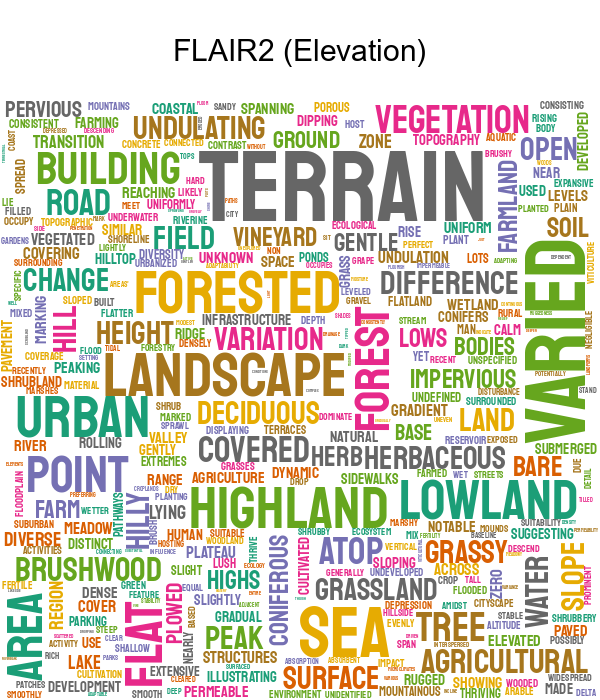}
    \caption{Word cloud for the Flair2-Elevation-caption dataset.}
    \label{fig:flair_elevation}
\end{figure}

\begin{figure}[tp]
    \centering
    \includegraphics[width=0.92\linewidth]{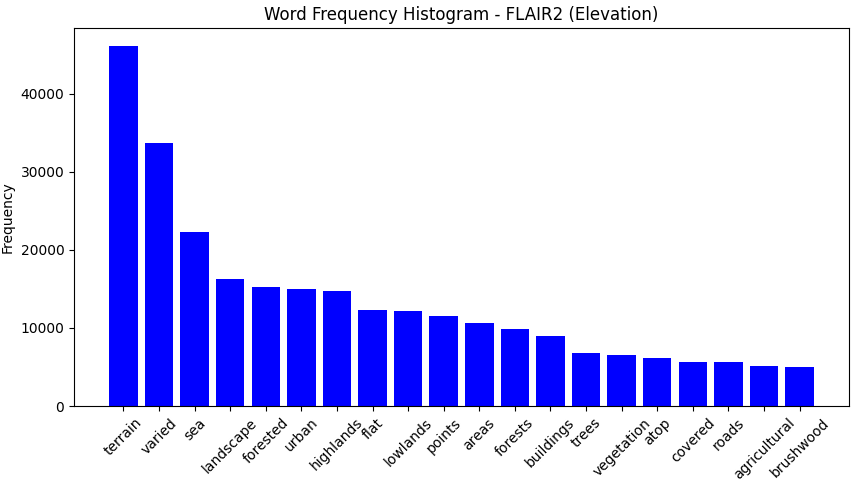}
    \caption{Histogram for the Flair2-elevation-caption dataset. Similarly to the other datasets of this appendix, FLAIR2-Elevation-caption has a focused vocabulary with many rare domain-specific terms. Its distribution deviates from a true power law due to frequent word repetition and an unusually long tail that is populated with remote sensing imagery terms. This mix of common and rare words could help models generalize by balancing domain-specific terms with diverse scene descriptors.}
    \label{fig:hist_flair_elevation}
\end{figure}
To begin, we select 61,711 images from the FLAIR~\#2 dataset, each accompanied by a segmentation map and an elevation map. The segmentation map provides land cover information, where each pixel corresponds to a specific land class, while the elevation map encodes height values representing terrain variation. Additionally, a class label mapping is used to associate numerical class IDs with their corresponding semantic labels, such as ``forest,'' ``urban area,'' or ``water body.'' This combination of datasets enables a comprehensive understanding of both land cover distribution and elevation patterns.

For each selected image, we analyze the elevation distribution by identifying the highest and lowest elevation values within the image patch. The corresponding land cover types at these extreme points are then determined using the segmentation map. This extraction process helps characterize the relationship between land cover and elevation, capturing how different landscapes correspond to varying altitude levels.

Using the extracted elevation and land cover information, we design a structured prompt to generate descriptive and natural textual captions. The prompt explicitly states the highest and lowest elevation values and their associated land cover types, followed by a request to summarize the overall terrain characteristics. The model is guided to describe whether the landscape is relatively flat or exhibits significant elevation changes, ensuring that the generated captions provide useful insights for downstream geospatial tasks.

To generate the final dataset, the structured prompts are input into a language model Pixtral~12B, which produces detailed, human-like descriptions of the elevation characteristics. These generated captions are then stored alongside their corresponding elevation and segmentation maps, forming the Flair2-Elevation-Caption dataset. The dataset offers rich, multimodal learning signals, allowing models to develop a deeper understanding of both land cover semantics and terrain variations. 

\begin{mdframed}[backgroundcolor=gray!5, linecolor=black]
This is an elevation map that indicates the height of each pixel.  
The highest areas, at an elevation of approximately [${highest\_height}$] meters, are [$highest\_class$].  
The lowest areas, at an elevation of approximately [${lowest\_height}$] meters, are [$lowest\_class$].  

Based on the provided context and elevation values, generate a concise and accurate description of the elevation map.  
Describe the image by briefly addressing:  

1) The highest and lowest land cover types.  

2) Whether the terrain is relatively flat or has significant elevation differences.
\end{mdframed}
In Fig. \ref{fig:flair_elevation}, we demonstrate the word cloud to provide a better overview of the dataset. The histogram of the Flair2-Elevation-caption dataset is presented in Fig. \ref{fig:hist_flair_elevation}.

\subsubsection{MMflood-SAR-caption}
The MMflood-SAR-caption dataset contains textual descriptions of flood maps generated using Pixtral~12B. In the following, we will describe the generation process in detail. Each flood map is represented as a binary mask, where white pixels indicate flooded areas, and black pixels represent non-flooded regions. To create meaningful textual descriptions, we analyze the extent of flooding and its spatial distribution before constructing structured prompts to guide the language model.
\begin{figure}[tp]
    \centering
    \includegraphics[width=0.92\linewidth]{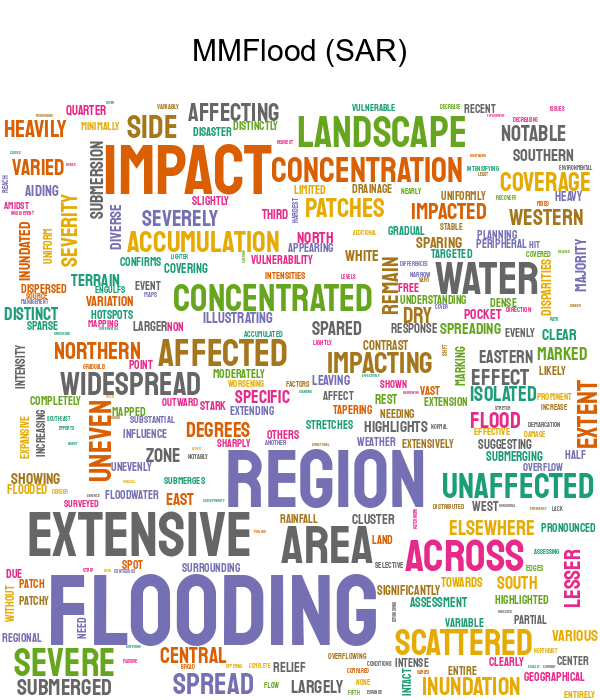}
    \caption{Word cloud for the MMflood-SAR-caption dataset.}
    \label{fig:mmflood_sar}
\end{figure}

\begin{figure}[tp]
    \centering
    \includegraphics[width=0.92\linewidth]{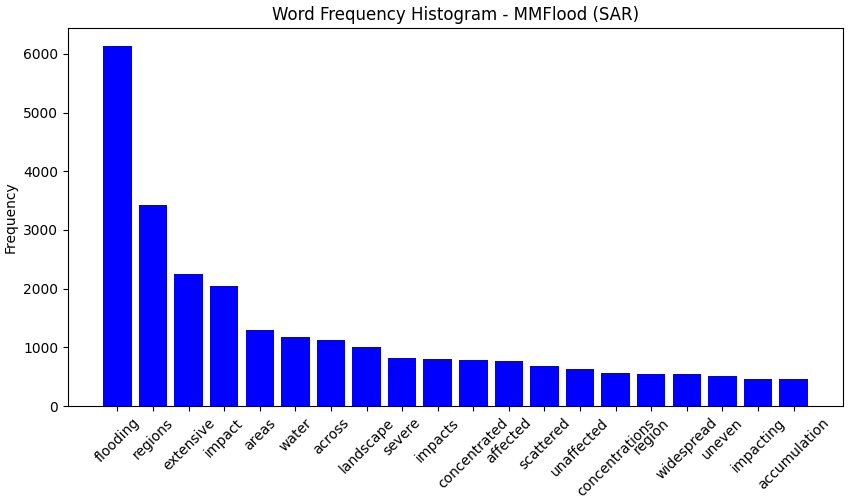}
    \caption{Word histogram for the MMflood-SAR-caption dataset.}
    \label{fig:hist_mmflood_sar}
\end{figure}
To quantify the severity of flooding, we calculate the total number of flooded pixels in each binary mask and express it as a percentage of the total image area. This value provides a direct measure of the flood extent, allowing the generated captions to capture variations in flood severity across different samples. In addition to flood coverage, we assess spatial patterns by dividing the image into four quadrants: top, bottom, left, and right. By checking the presence of flooded pixels in each region, we determine whether the flooding is concentrated in specific areas or dispersed across the map.

Once these characteristics are extracted, we construct a structured prompt to guide Pixtral~12B in generating concise and informative descriptions. The prompt instructs the model to summarize the proportion of flooded areas and specify their locations within the image. To ensure clarity and brevity, the generated description is limited to fewer than 70 words, making it suitable for vision--language pretraining.

The prompts are then fed into Pixtral~12B, which generates natural language captions describing each flood map. These captions, stored alongside their corresponding binary masks, form the MMFlood dataset, enabling automated flood analysis, disaster monitoring, and geospatial AI research. By combining semantic flood mapping with natural language descriptions, this dataset enhances the ability of AI models to interpret and describe flood-affected regions in a human-readable format. We also provide the word cloud in Fig.~\ref{fig:mmflood_sar} to show the detailed content of this dataset. In Fig.~\ref{fig:hist_mmflood_sar}, we present the histogram of the words in the MMflood-SAR-caption dataset.

\begin{mdframed}[backgroundcolor=gray!5, linecolor=black] This is a flood map where areas marked with white pixels indicate flooded regions.
The flooded area occupies approximately [$flood\_percentage$]\% of the entire map.
Please analyze the flood map and provide insights into the affected areas.

You must generate a short description (less than 70 words) of the elevation image.
First, describe the portion of floods; then introduce the location of the flooded areas. \end{mdframed}

After generating these detailed textual descriptions, we further refine the dataset by prompting Pixtral~12B to summarize each description into a concise 30-word caption. The prompts used in this summarizing process are as follows:

\begin{mdframed}[backgroundcolor=gray!5, linecolor=black] You are an AI assistant tasked with creating a concise 30-word caption that effectively summarizes the key points of the following content:
[$caption$] \end{mdframed}

\subsubsection{NLCD-hyper-caption}
The caption generation process for hyperspectral images follows a methodology similar to the FLAIR~\#2 dataset, ensuring consistency in semantic information extraction and description generation. Hyperspectral images provide rich spectral information across multiple wavelengths, allowing for a more detailed classification of land cover types. However, the caption generation process remains aligned with the Flair2-RGB-Caption dataset, differing only in the specific prompt format used to guide the Pixtral~12B model.
\begin{figure}[tp]
    \centering
    \includegraphics[width=0.92\linewidth]{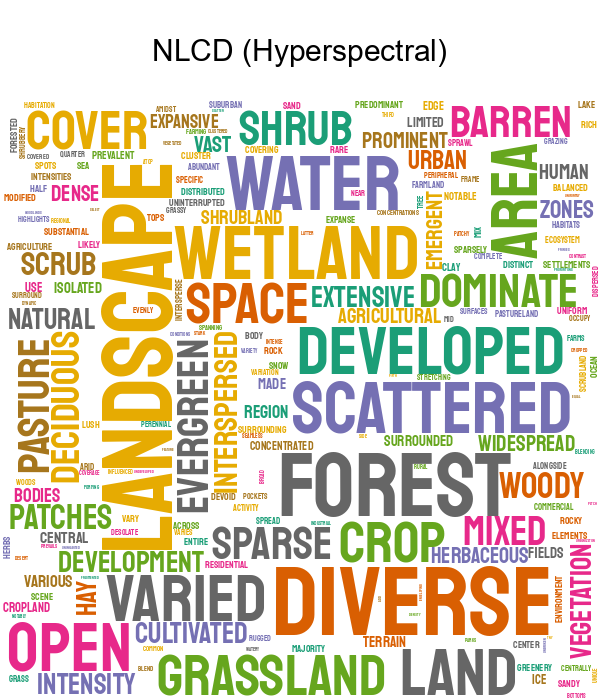}
    \caption{Word cloud for the NLCD-hyper-caption dataset.}
    \label{fig:nlcd_hyper}
\end{figure}
\begin{figure}[tp]
    \centering
    \includegraphics[width=\linewidth]{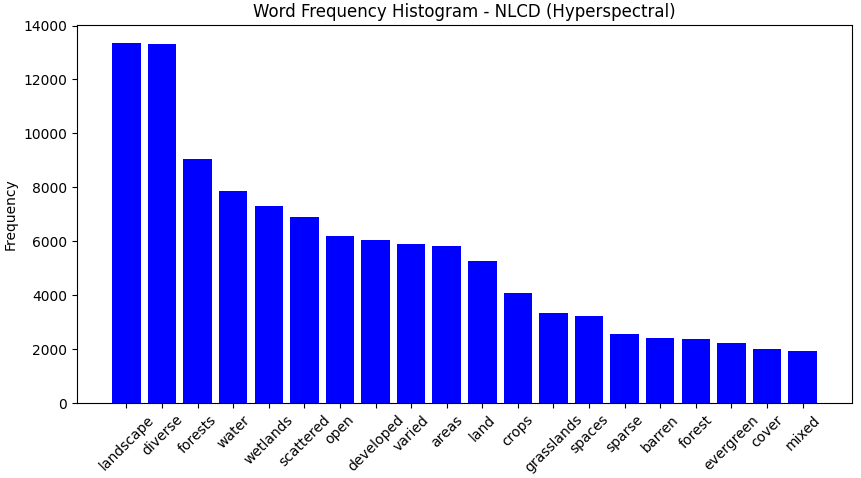}
    \caption{Word histogram for the NLCD-hyper-caption dataset.}
    \label{fig:hist_nlcd_hyper}
\end{figure}
To create captions, we first extract semantic information from the hyperspectral images using their corresponding segmentation maps. Each land cover type is identified and labeled, and its spatial distribution within the image is analyzed. Unlike conventional RGB imagery, where color-based segmentation might play a role, hyperspectral images rely on spectral signatures to differentiate land cover types, making this process crucial for accurate caption generation.

Once the land cover labels and their distributions are determined, we construct a structured prompt to generate natural language descriptions. The Pixtral~12B model is tasked with producing concise and human-readable descriptions of the scene. The prompt ensures that the generated captions focus solely on land cover types and their spatial distributions, avoiding any mention of technical details, color coding, or hyperspectral data intricacies.

Following the generation of detailed textual descriptions, we further prompt Pixtral~12B to refine the descriptions into brief and natural captions. This step ensures that the dataset provides both long-form and short-form textual annotations, improving its usability for vision--language tasks, remote sensing analysis, and multimodal learning.

The prompt template used for hyperspectral image captioning is as follows:

\begin{mdframed}[backgroundcolor=gray!5, linecolor=black] 
You are an AI visual assistant capable of describing a scene based on a segmentation map. The map represents different land cover types using specific colors. The legend is as follows:

- [$color\_i$] corresponds to [$label\_i$], occupying [$percent\_i$]\% of the image.

Do not mention any colors, color coding, or technical details.
Use the given class names. Only mention land cover types in the color legend.

Generate a brief and natural description of the scene by refining:
``The hyperspectral image contains [$presented\_labels$] land types.''
Provide a concise description of their spatial distributions (e.g., left, right, top, bottom). \end{mdframed}

To provide a clearer overview of this dataset, we showcase the word cloud in Fig.~\ref{fig:nlcd_hyper}. We also showcase the histogram of the words in the NLCD-hyper-caption dataset in Fig.~\ref{fig:hist_nlcd_hyper}.

\subsubsection{SAR-ship-caption and IR-ship-caption}
For the SAR-ship-caption and IR-ship-caption subsets, we derive short captions from existing datasets to ensure high-quality textual descriptions. For IR-ship-caption, we utilize samples from the HIT-UAV dataset, specifically the data provided in \citet{zhang2024earthgpt}, which is originally designed for visual question answering tasks. To generate descriptions for infrared images, we concatenate the corresponding question and answer pairs, transforming them into concise yet informative captions.

Similarly, for the SAR-ship-caption dataset, we generate captions for SAR images using three key datasets: HRSID~\cite{wei2020hrsid}, SSDD~\cite{zhang2021sar}, and AIR-SARShip-2.0~\cite{sun2020high}. These datasets provide extensive annotated SAR imagery, enabling the creation of meaningful captions. For further details on the captioning methodology, please refer to \citet{zhang2024earthgpt}.

\subsubsection{ChatEarthNet captions}
\begin{figure}[tp]
    \centering
    \includegraphics[width=0.92\linewidth]{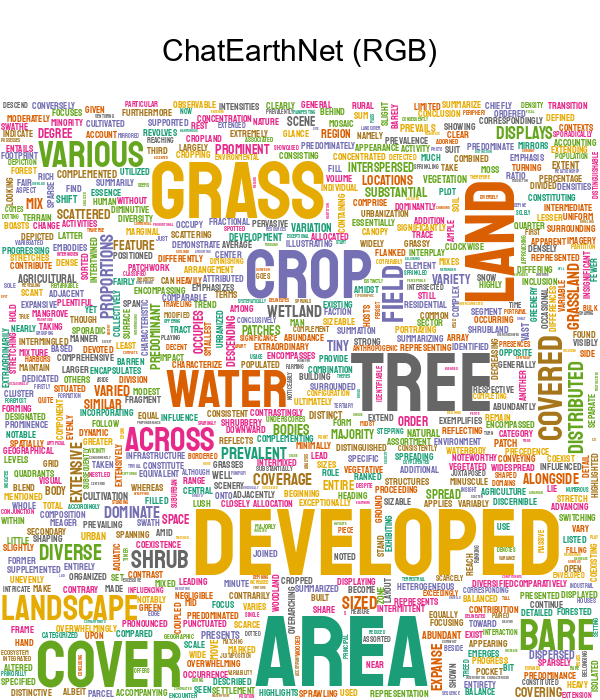}
    \caption{Word cloud for the ChatEarthNet-caption datasets.}
    \label{fig:cen}
\end{figure}

\begin{figure}[tp]
    \centering
    \includegraphics[width=0.92\linewidth]{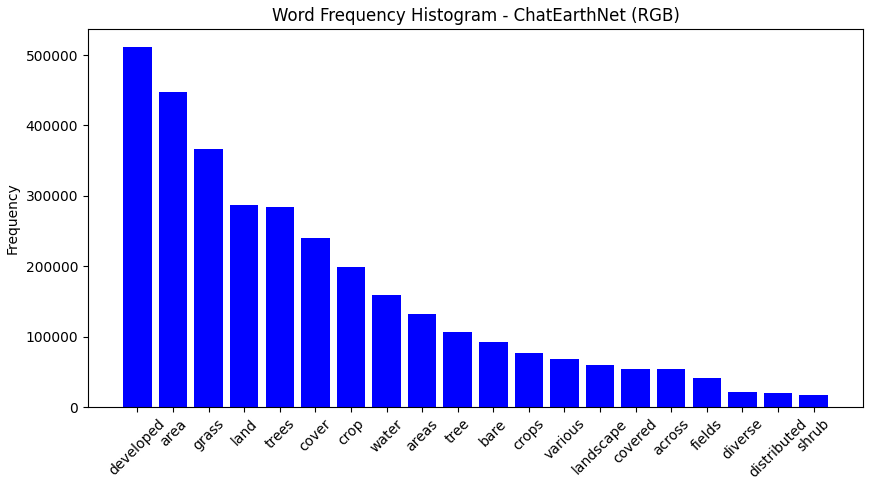}
    \caption{Word histogram for the ChatEarthNet-caption datasets.}
    \label{fig:hist_cen}
\end{figure}
Regarding the ChatEarthNet-S2-caption and ChatEarthNet-SAR-caption datasets, we reuse the detailed descriptions provided in the original ChatEarthNet dataset. Specifically, we use Pixtral~12B~\cite{agrawal2024pixtral} to summarize these detailed descriptions into short image captions and generate question-answer pairs for visual question answering tasks. The word cloud for ChatEarthNet-caption datasets is presented in Fig.~\ref{fig:cen}. To provide more insights into the dataset, we also show the histogram in Fig.~\ref{fig:hist_cen}. The prompt is as follows:
\begin{mdframed}[backgroundcolor=gray!5, linecolor=black] You are an AI assistant tasked with creating a concise 30-word caption that effectively summarizes the key points of the following content:
[$caption$] \end{mdframed}

\subsection{More dataset examples}
These paired samples not only capture scene-level descriptions but also highlight fine-grained properties, such as object positions and fine-grained land cover types. Thereby the dataset enables the development of \emph{foundation models} proficient in downstream performance for land cover mapping, flood monitoring, and object detection under challenging conditions (e.g., night or cloudy weather). 

\begin{figure*}[tp]
    \centering
    \includegraphics[width=0.85\linewidth]{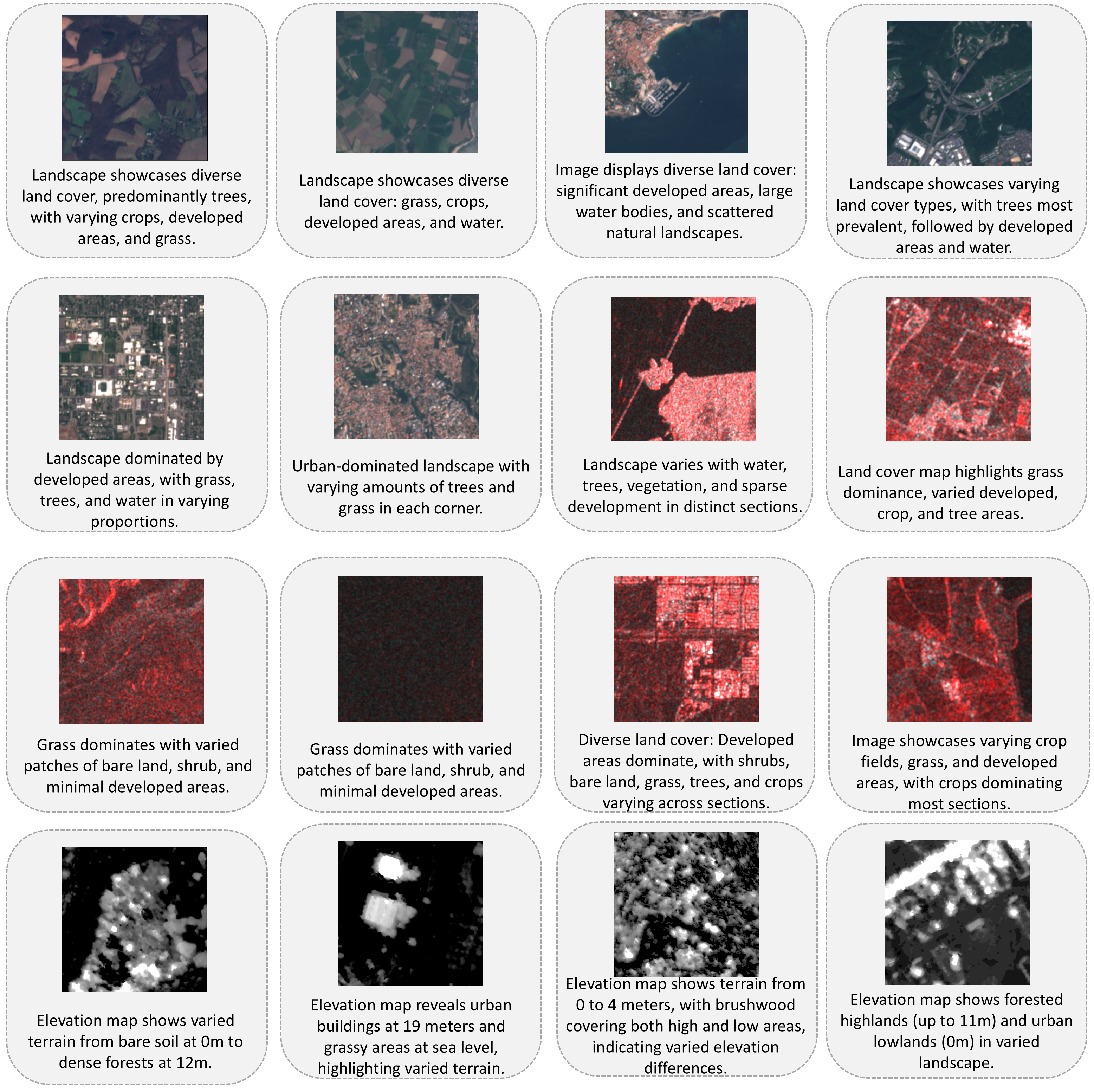}
    \caption{More examples of the image-text pair of GeoLangBind-2M, our multimodal image-text dataset.}
    \label{fig:example1}
\end{figure*}

\begin{figure*}[tp]
    \centering
    \includegraphics[width=0.85\linewidth]{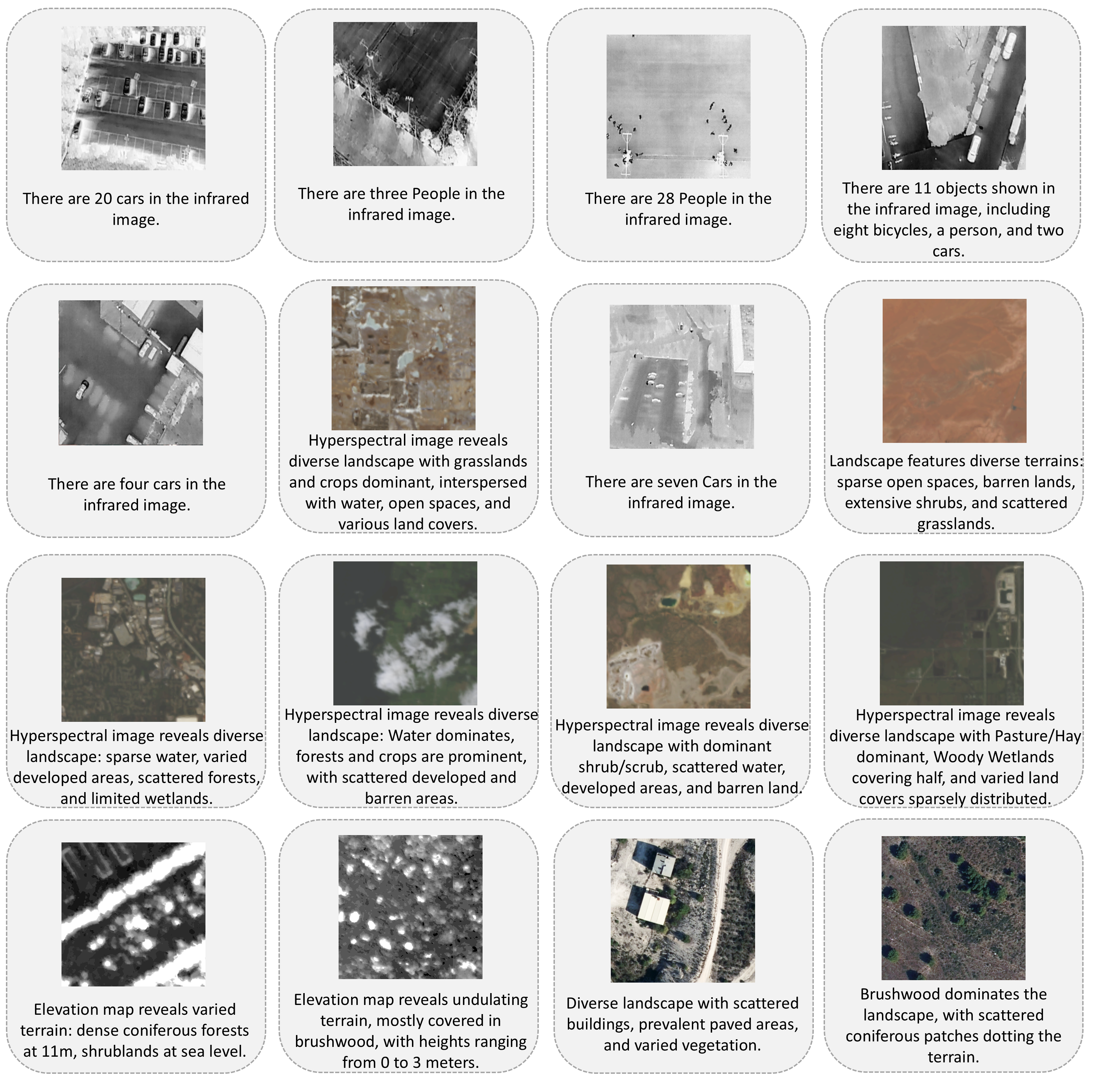}
    \caption{More examples of the image-text pair of GeoLangBind-2M, our multimodal image-text dataset.}
    \label{fig:example2}
\end{figure*}

In Fig.~\ref{fig:example1} and Fig.~\ref{fig:example2}, we provide more examples to provide a better overview of our dataset.

\subsection{Ablation studies on distillation loss terms}

\begin{table}[ht]
    \centering
    \caption{Distillation loss balancing weights vs.\ accuracy}
    \scalebox{0.75}
    {
    \begin{tabular}{ccccc}
    \toprule
    $\alpha_s$ (SigLIP) & $\alpha_d$ (DinoV2) & $\alpha_v$ (ViT) & aid (Top-1)  & eurosat (Top-1)  \\
    \midrule
    0 & 0 & 0 & 59.10 & 32.96 \\
    1.0 & 0.0 & 0.0 & 61.42 & 33.63 \\
    1.0 & 1.0 & 0.0 & 65.36 & 35.26 \\
    1.0 & 1.0 & 1.0 & \underline{68.75} & \textbf{38.63} \\
    2.0 & 0.0 & 0.0 & 64.85 & 33.19 \\
    2.0 & 0.5 & 0.5 & 64.70 & 25.30 \\
    \rowcolor{gray!25} 2.0 & 1.0 & 1.0 & \textbf{69.20} & \underline{36.67} \\
    \bottomrule
    \end{tabular}
    }
    \label{tab:ratios}
\end{table}
Table~\ref{tab:ratios} presents an ablation study on the impact of different distillation loss balancing weights (\(\alpha_s\), \(\alpha_d\), \(\alpha_v\)) on classification accuracy for the AID and EuroSAT datasets. The baseline model (no distillation) achieves 59.10\% (AID) and 32.96\% (EuroSAT). Adding SigLIP distillation (\(\alpha_s = 1.0\)) improves performance, while combining SigLIP and DinoV2 (\(\alpha_s = 1.0, \alpha_d = 1.0\)) further enhances results, especially for AID (65.36\%). Introducing ViT distillation (\(\alpha_v = 1.0\)) boosts EuroSAT accuracy, peaking at 38.63\% when \(\alpha_s = \alpha_d = \alpha_v = 1.0\). In this work, we adopt the best overall configuration, \(\alpha_s = 2.0, \alpha_d = 1.0, \alpha_v = 1.0\), achieves the highest AID accuracy (69.20\%) and strong EuroSAT performance (36.67\%). This demonstrates the effectiveness of multi-teacher distillation.

\subsection{Confusion matrices of zero-shot classification tasks}

\begin{figure}[tp]
    \centering
    \includegraphics[width=\linewidth]{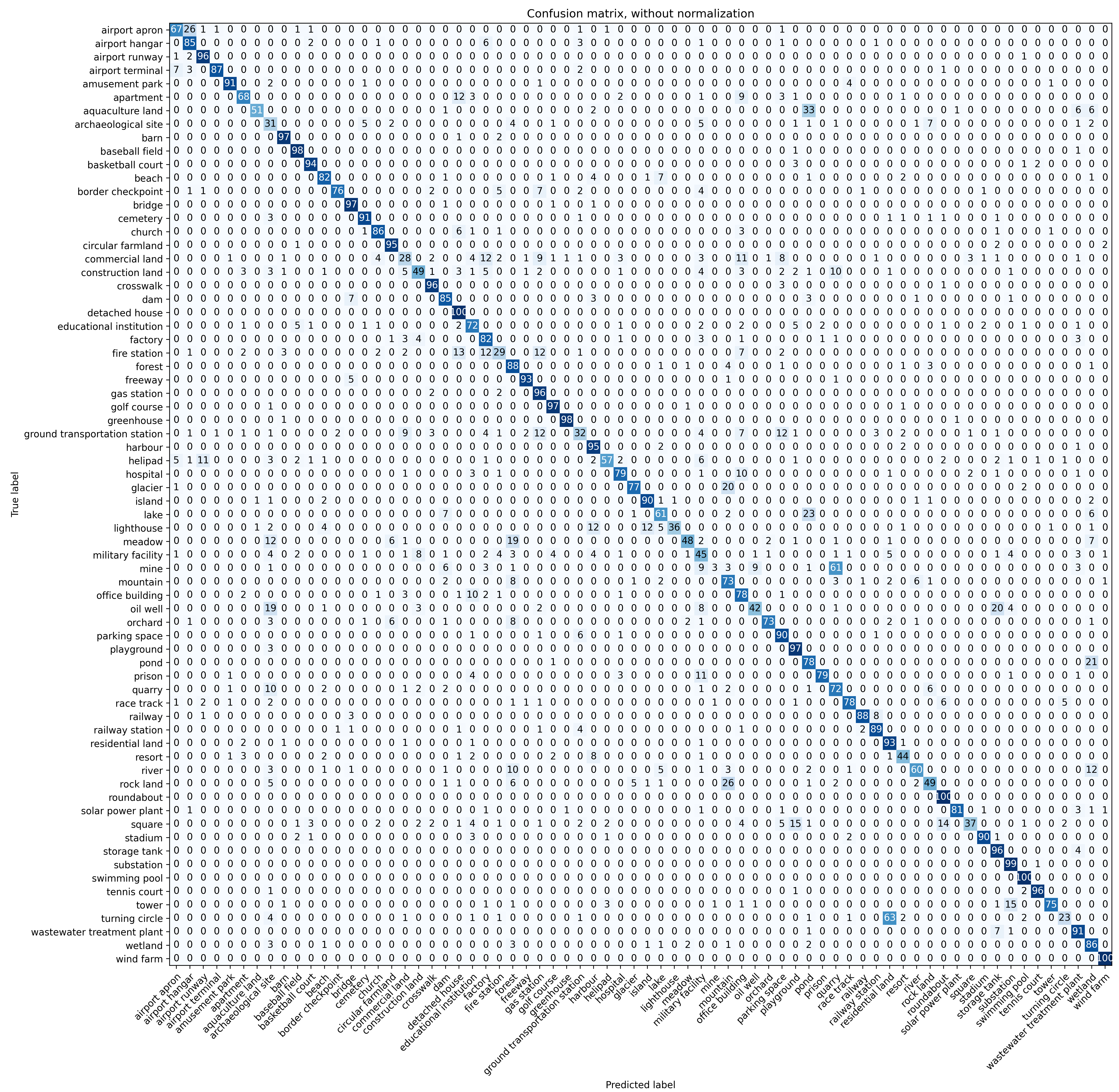}
    \caption{Confusion matrix of the zero-shot classification results of DOFA-CLIP-L-384 on SkyScript dataset.}
    \label{fig:skyscript}
\end{figure}

\begin{figure}[tp]
    \centering
    \includegraphics[width=\linewidth]{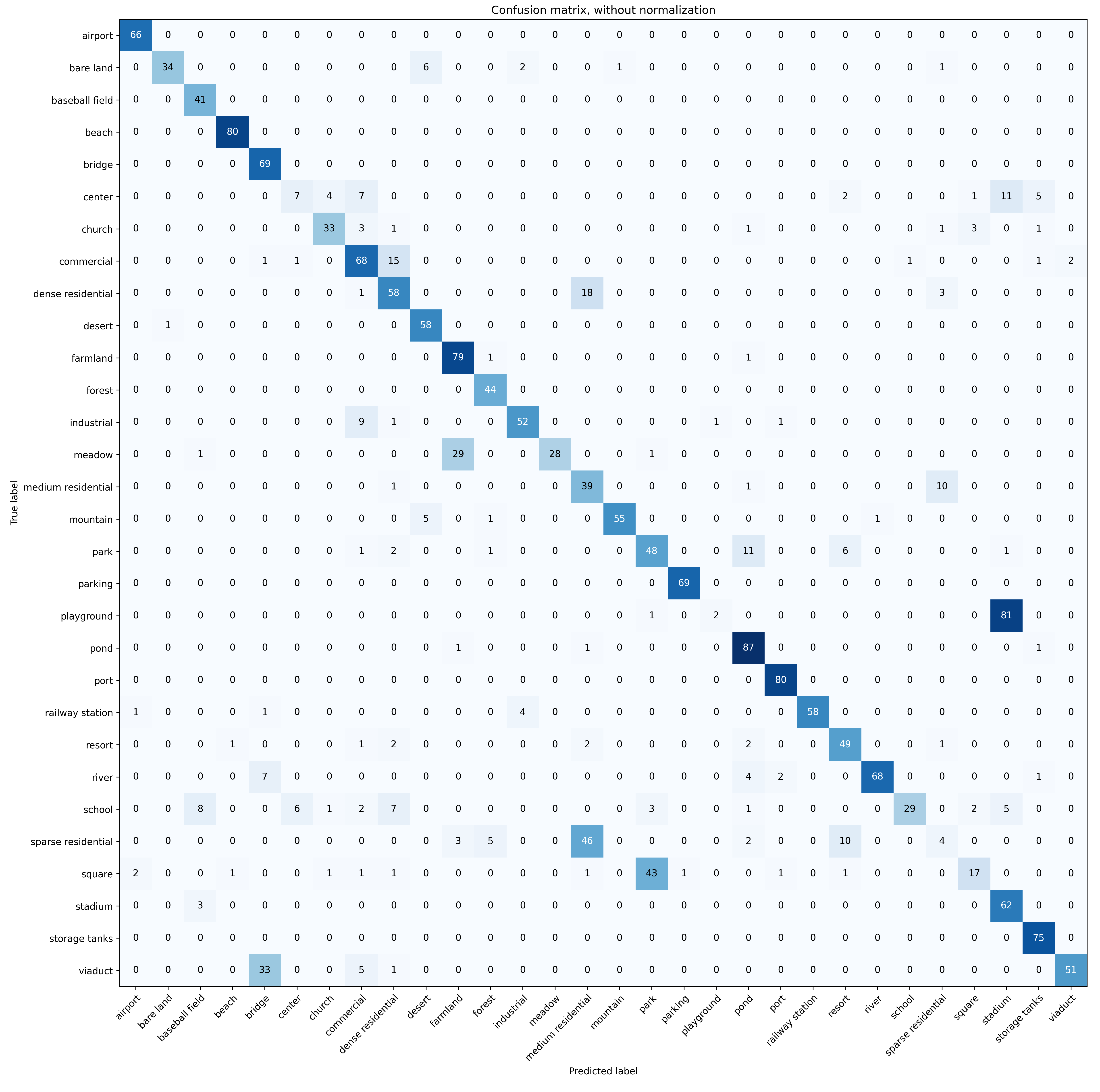}
    \caption{Confusion matrix of the zero-shot classification results of DOFA-CLIP-L-384 on AID dataset.}
    \label{fig:aid}
\end{figure}

\begin{figure}[tp]
    \centering
    \includegraphics[width=\linewidth]{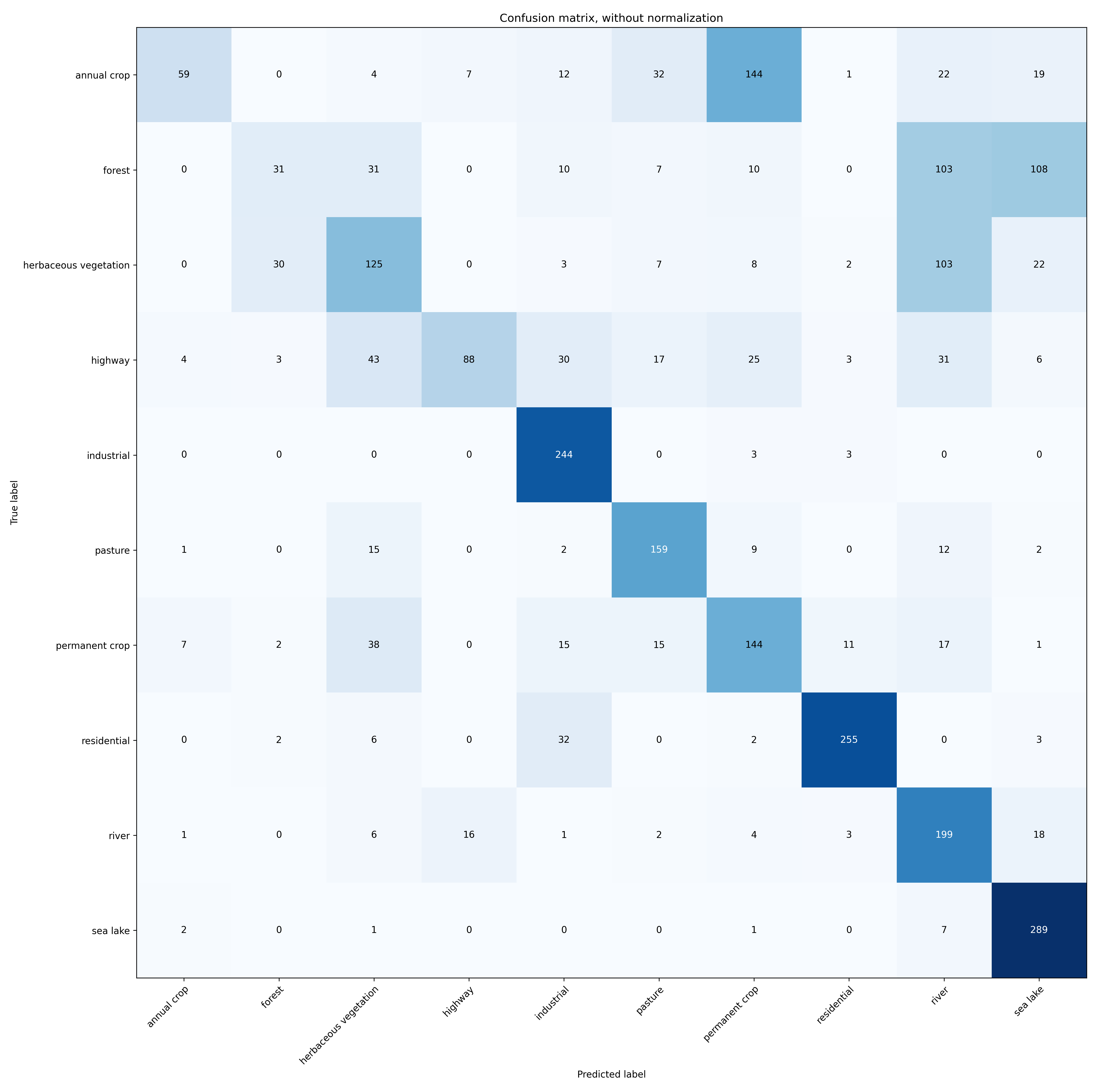}
    \caption{Confusion matrix of the zero-shot classification results of DOFA-CLIP-L-384 on EuroSAT dataset.}
    \label{fig:eurosat}
\end{figure}

\begin{figure}[tp]
    \centering
    \includegraphics[width=\linewidth]{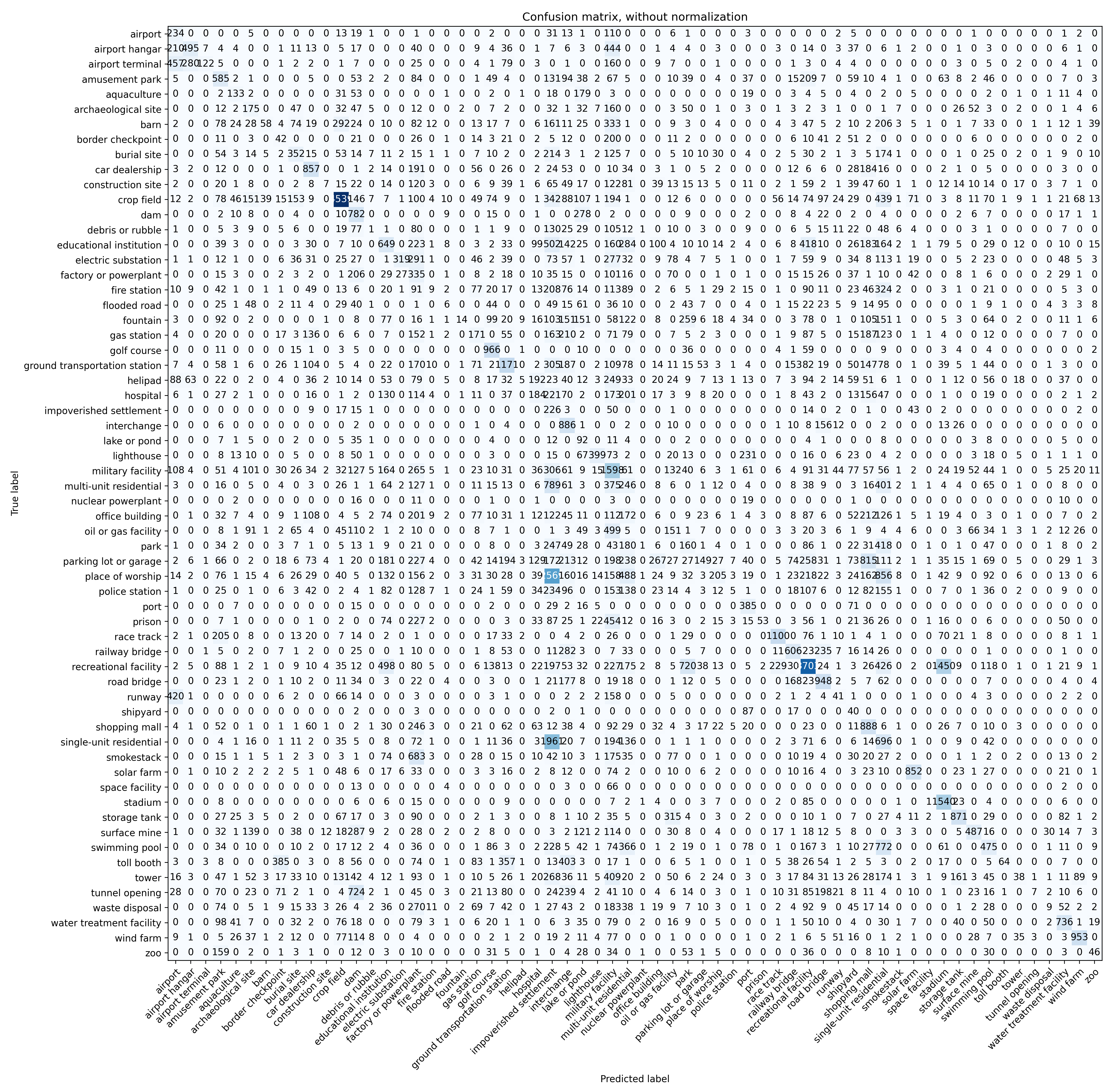}
    \caption{Confusion matrix of the zero-shot classification results of DOFA-CLIP-L-384 on fMoW dataset.}
    \label{fig:fmow}
\end{figure}

\begin{figure}[tp]
    \centering
    \includegraphics[width=\linewidth]{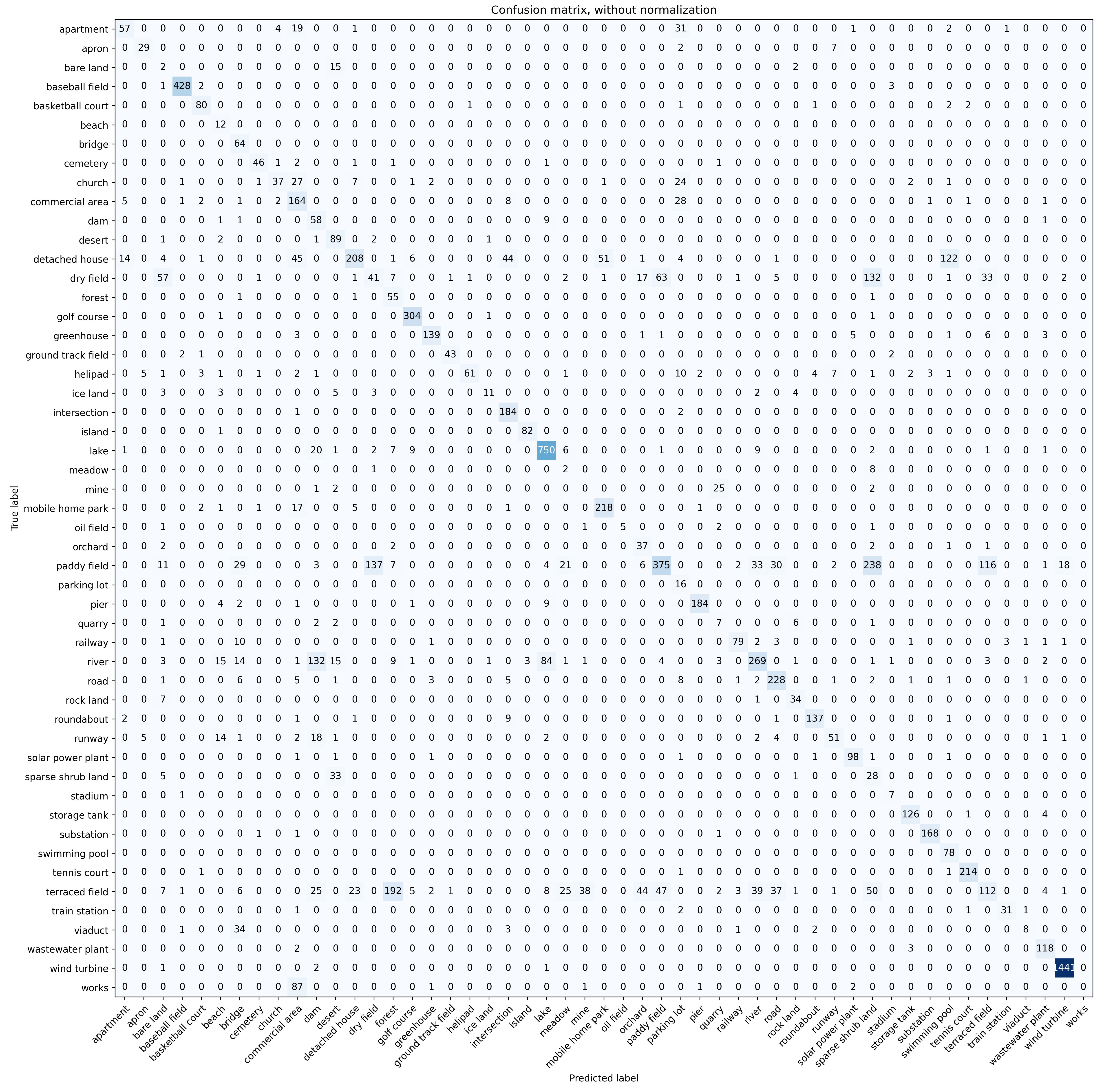}
    \caption{Confusion matrix of the zero-shot classification results of DOFA-CLIP-L-384 on Million-AID dataset.}
    \label{fig:millionaid}
\end{figure}

\begin{figure}[tp]
    \centering
    \includegraphics[width=\linewidth]{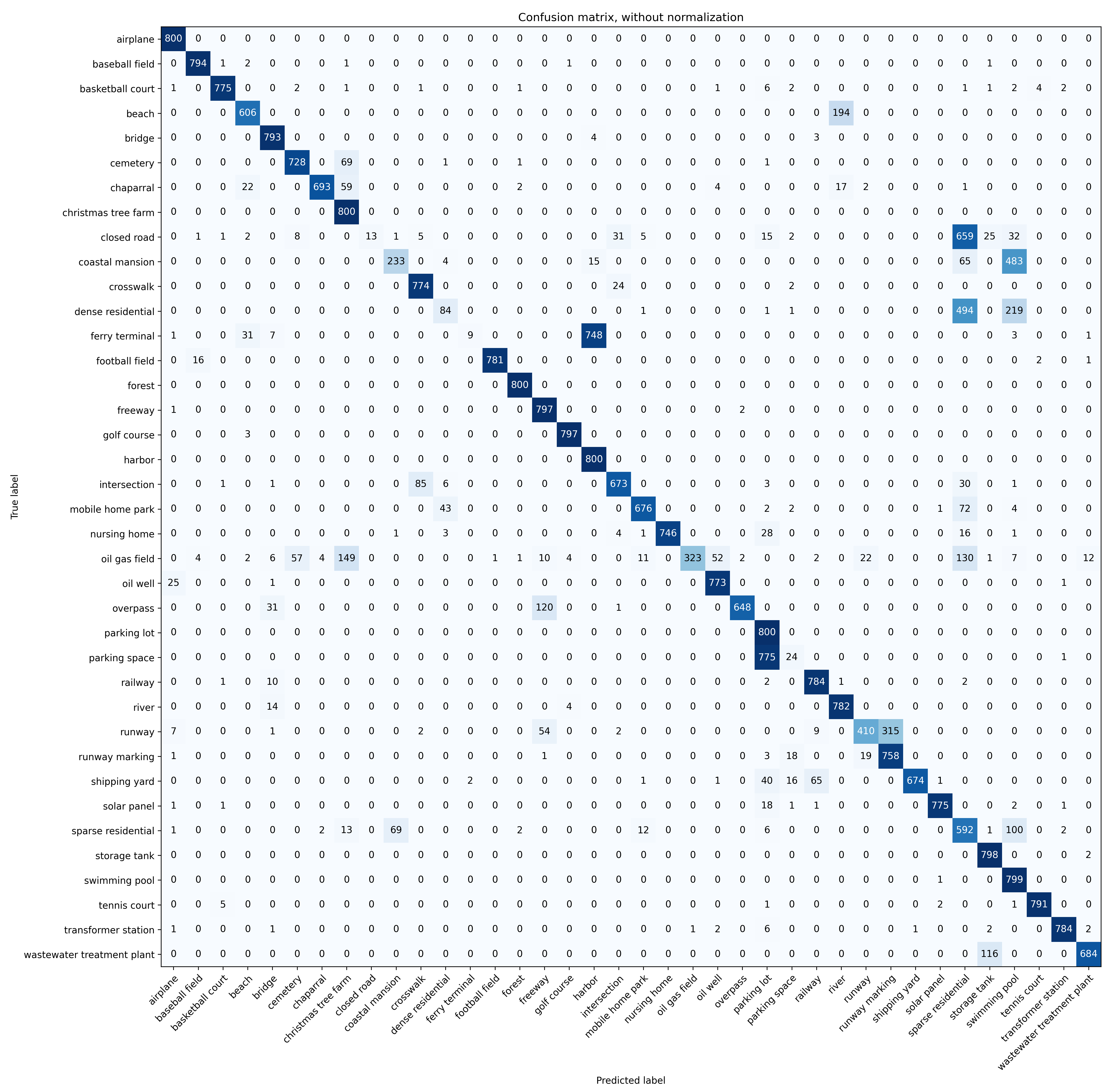}
    \caption{Confusion matrix of the zero-shot classification results of DOFA-CLIP-L-384 on PatternNet dataset.}
    \label{fig:patternnet}
\end{figure}

\begin{figure}[tp]
    \centering
    \includegraphics[width=\linewidth]{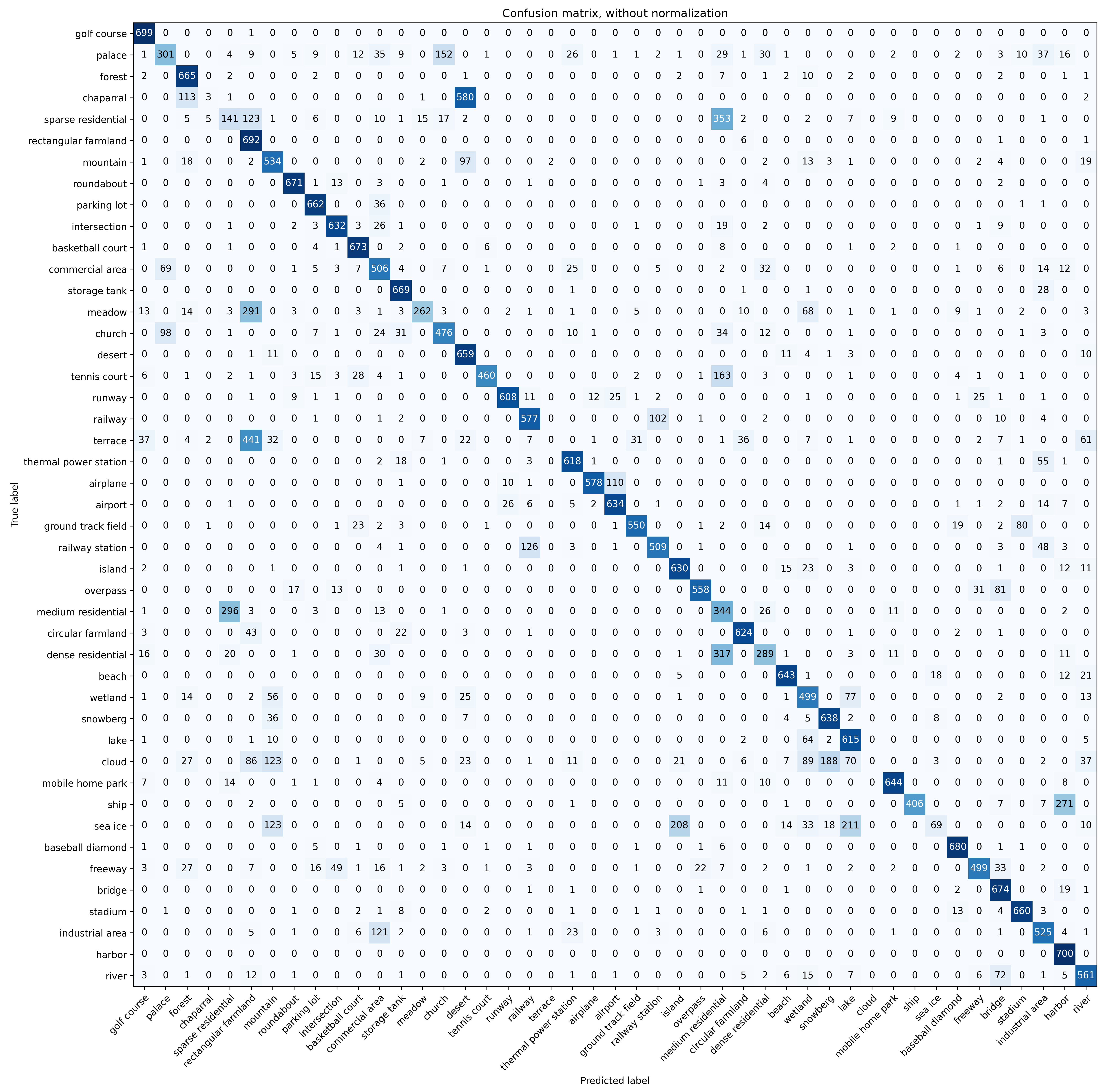}
    \caption{Confusion matrix of the zero-shot classification results of DOFA-CLIP-L-384 on the RESISC45 dataset.}
    \label{fig:nwpu}
\end{figure}

\begin{figure}[tp]
    \centering
    \includegraphics[width=\linewidth]{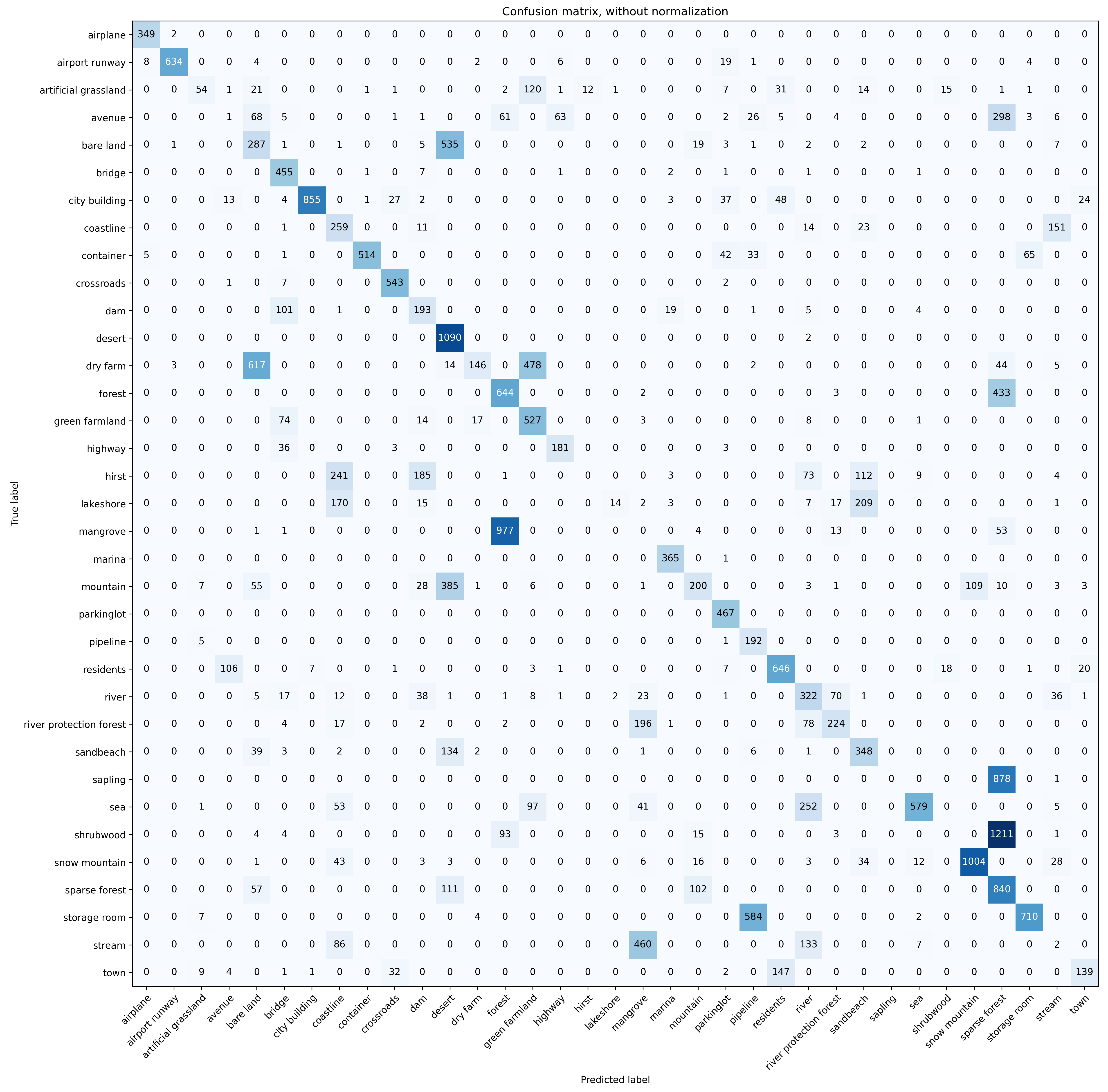}
    \caption{Confusion matrix of the zero-shot classification results of DOFA-CLIP-L-384 on RSICB dataset.}
    \label{fig:rsicb}
\end{figure}

Figure~\ref{fig:skyscript} to Fig.~\ref{fig:rsicb} present the confusion matrices of our DOFA-CLIP-L-384 model on all eight datasets for zero-shot classification. These confusion matrices reveal that common model mistakes largely occur for classes that humans would also have trouble distinguishing, or for classes with similar semantic meanings. For instance, the model commonly confuses ``sparse residential'' and ``medium residential'' (RESISC45), ``annual crop'' and ``permanent crop'' (EuroSAT), ``aquaculture land'' and ``pond'' (SkyScript), ``chaparral'' and ``desert'' (RESISC45), ``mine'' and ``quarry'' (SkyScript, Million-AID), ``recreational facility'' and ``stadium'' and ``playground'' (fMoW, AID), and ``ship'' and ``harbor'' and ``ferry terminal'' (RESISC45, PatternNet).

\begin{figure*}[tp]
    \centering
    \includegraphics[width=\linewidth]{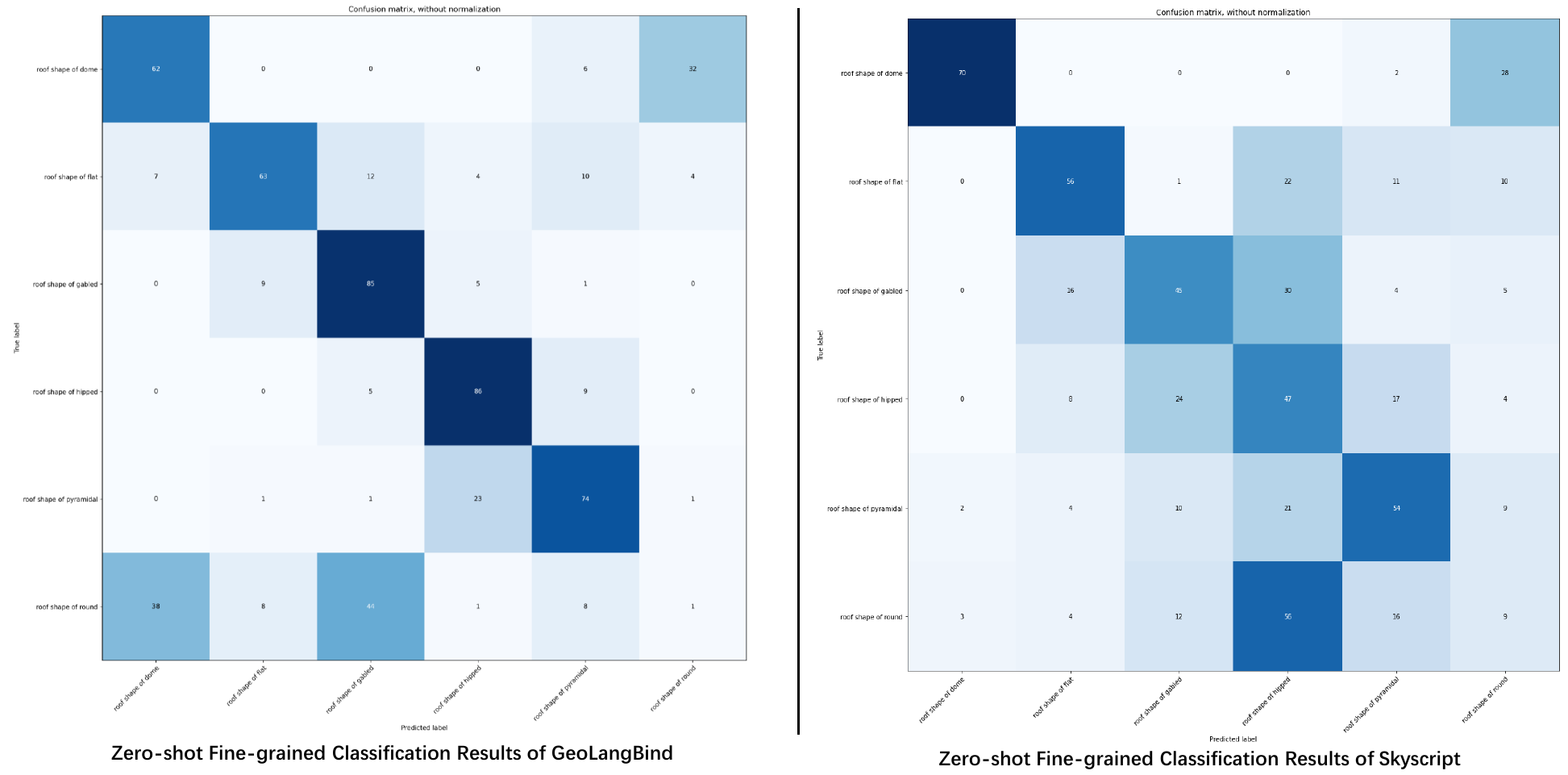}
    \caption{Comparison of confusion matrices between SkyScript-Large and DOFA-CLIP-L-384 on the fine-grained zero-shot classification task.}
    \label{fig:cmcompare}
\end{figure*}

In Fig.~\ref{fig:cmcompare}, we show the comparison of confusion matrices between SkyScript (ViT-L) and DOFA-CLIP-L-384 on fine-grained zero-shot classification tasks. It can be clearly seen that our model performs better. 

\subsection{More visualization of features}
\begin{figure*}[tp]
    \centering
    \includegraphics[width=\linewidth]{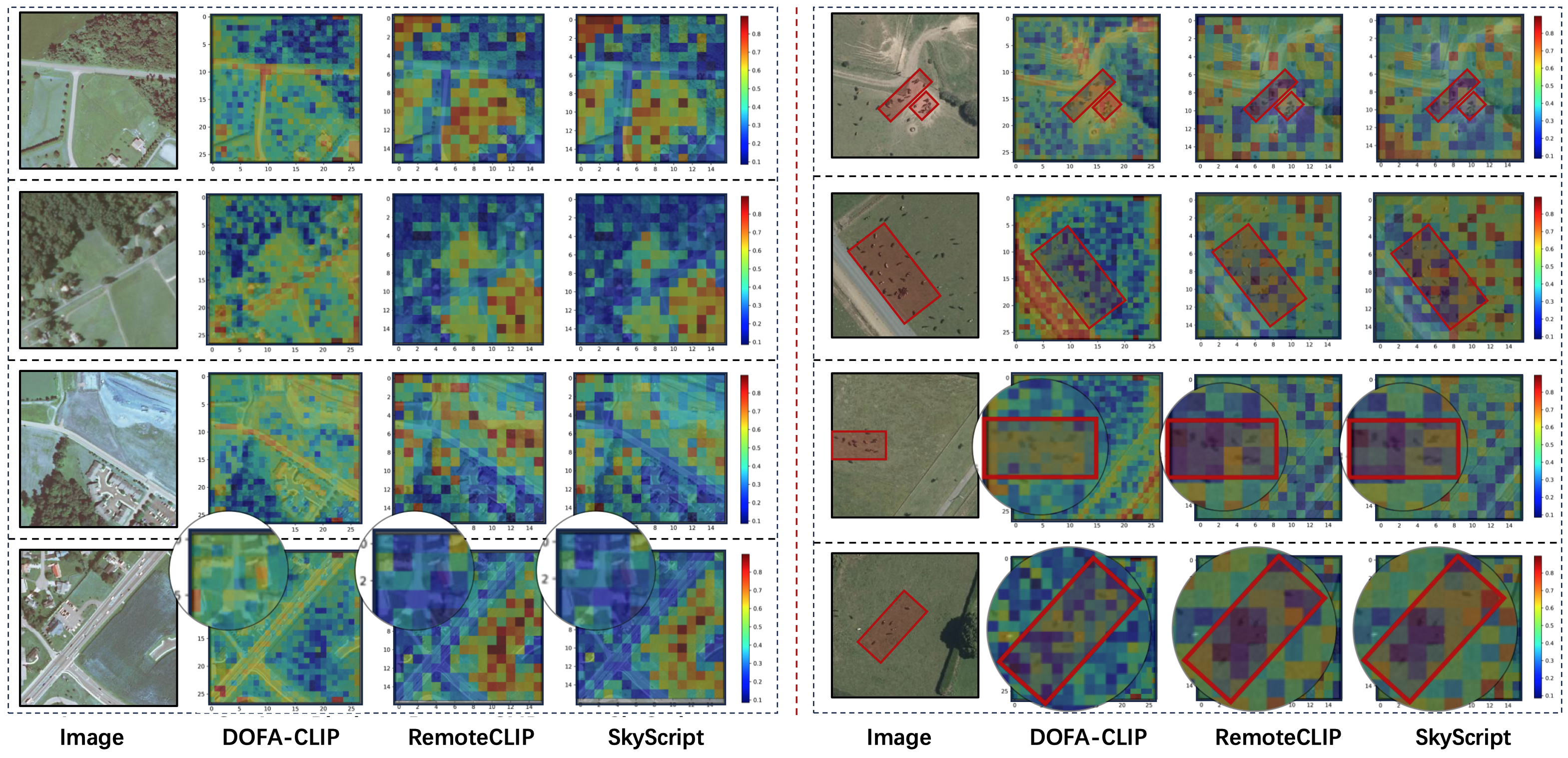}
    \caption{Visualization of features from RemoteCLIP (ViT-L), SkyScript (ViT-L), and DOFA-CLIP-L-384 on the m-chesapeake and m-nz-cattle datasets from GEO-Bench.}
    \label{fig:supp_vis_feat}
\end{figure*}

To better understand the representations learned by different models, Fig. \ref{fig:supp_vis_feat} visualizes the feature maps extracted from RemoteCLIP (ViT-L), SkyScript (ViT-L), and DOFA-CLIP-L-384 across two datasets: m-chesapeake and m-nz-cattle from GEO-Bench. The m-chesapeake dataset primarily focuses on land cover classification, while the m-nz-cattle dataset captures cattle in remote sensing images. The feature visualization highlights key differences in spatial structure, consistency, and detail preservation across models.

\paragraph{Spatially structured features} DOFA-CLIP produces features with higher spatial resolution, enabling finer spatial distinctions compared to RemoteCLIP and SkyScript. The feature maps exhibit well-defined structures that align with meaningful spatial regions in the original images, such as roads, buildings, and vegetation. 

\paragraph{Consistency across semantic classes} The feature maps from DOFA-CLIP demonstrate greater consistency across images with similar semantic content. As observed, buildings and roads consistently exhibit high activations, while tree-covered areas show lower activations. This structured activation pattern is apparent across multiple samples, suggesting that DOFA-CLIP learns more stable and semantically aligned representations.

\paragraph{Preservation of details} DOFA-CLIP retains finer details within its feature maps, capturing small-scale variations and object boundaries more effectively than the baselines. This advantage is particularly evident in the m-nz-cattle dataset, where cattle are localized with sharper feature activations, as highlighted by the red bounding boxes. RemoteCLIP and SkyScript, on the other hand, exhibit more diffused and less distinct feature responses, making object-level distinctions less precise.

\bibliography{aaai2026}

\end{document}